\documentclass[compsoc, conference, a4paper, 10pt, times]{IEEEtran}

\usepackage{amsmath,amssymb,amsfonts}
\usepackage{algorithmic}
\usepackage{graphicx}
\usepackage{textcomp}
\usepackage[table,xcdraw]{xcolor}
\usepackage{booktabs}
\usepackage[hidelinks]{hyperref}

\usepackage{multirow}
\usepackage[backend=bibtex, style=numeric, minnames=1, maxnames=1, maxcitenames=1, sorting=none]{biblatex}
\usepackage{wasysym}
\usepackage{xspace}

\newcommand{\mypara}[1]{\vspace{5pt}\noindent{\textbf{{#1}}\xspace}}
\newcommand{\myparait}[1]{\vspace{5pt}\noindent{\textit{{{#1}}}\xspace}}

\definecolor{framecolor}{HTML}{E3E3E3}

\usepackage[framemethod=tikz]{mdframed}
\mdfdefinestyle{remarkstyle}{
backgroundcolor=framecolor,
roundcorner=2pt,
innerleftmargin=2pt,
innerrightmargin=2pt,
innertopmargin=2pt,
innerbottommargin=2pt,
linewidth=0.5pt,
}

\newcounter{summ}[section]

\newcounter{takeawyacount}
\begin{document}

\title{David and Goliath: An Empirical Evaluation of Attacks and Defenses\\for QNNs at the Deep Edge}


\author{\IEEEauthorblockN{1\textsuperscript{st} Miguel Costa}
\IEEEauthorblockA{\textit{ALGORITMI Research Centre} \\
Universidade do Minho, Portugal \\
miguel.costa@dei.uminho.pt}
\and
\IEEEauthorblockN{2\textsuperscript{nd} Sandro Pinto}
\IEEEauthorblockA{\textit{ALGORITMI Research Centre} \\
Universidade do Minho, Portugal \\
sandro.pinto@dei.uminho.pt}
}

\maketitle

\begin{abstract}
Machine learning (ML) is shifting from the cloud to the edge. Edge computing reduces the surface exposing private data and enables reliable throughput guarantees in real-time applications. Of the panoply of devices deployed at the edge, resource-constrained microcontrollers (MCUs), e.g., Arm Cortex-M, are more prevalent, orders of magnitude cheaper, and less power-hungry than application processors (APUs) or graphical processing units (GPUs). Thus, enabling intelligence at the so-called deep/extreme edge is the zeitgeist, with researchers focusing on unveiling novel approaches to deploy artificial neural networks (ANN) on these constrained devices. Quantization is a well-established technique that has proved effective, i.e., negligible impact on accuracy, in enabling the deployment of neural networks on MCUs; however, it is still an open question to understand the robustness of quantized neural networks (QNNs) in the face of well-known adversarial examples.

To fill this gap, we empirically evaluate the effectiveness of attacks and defenses from (full-precision) ANNs on (constrained) QNNs. Our evaluation suite includes three QNNs targeting TinyML applications, ten attacks, and six defenses. With this study, we draw a set of interesting findings. First, quantization increases the point distance to the decision boundary and leads the gradient estimated by some attacks to explode or vanish. Second, quantization can act as a noise attenuator or amplifier, depending on the noise magnitude, and causes gradient misalignment. Regarding adversarial defenses, we conclude that input pre-processing defenses show impressive results on small perturbations; however, they fall short as the perturbation increases. At the same time, train-based defenses increase adversarial robustness by increasing the average point distance to the decision boundary, which holds even after quantization. However, we argue that train-based defenses still need to smooth the quantization-shift and gradient misalignment phenomenons to counteract adversarial example transferability to QNNs. All artifacts are open-sourced to enable independent validation of results and encourage further exploration of the robustness of QNNs.
\end{abstract}



\section{Introduction}
The availability of large volumes of data and tremendous computing power turned machine learning (ML) into a critical technology to drive innovation and assist humans in tasks ranging from healthcare \cite{miotto2017} and finance \cite{ahmed2022} to mobility \cite{ying2021}. As ML workloads tend to be computationally intensive, most ML systems developed to date followed a cloud-centered design, sending the data collected at the edge to the cloud \cite{costa2022, murshed2021, stoica2017}. The cloud performs the ML computation and returns the final decision to the edge responsible for actuating over the environment. However, recent concerns about privacy, latency, power consumption, and cost are forcing ML to move from the cloud to the edge near the data source \cite{bernhard2019, gorsline2021}.

Shifting ML workloads to the edge reduces the exposure of private data, as there is no data transmission over the network \cite{bernhard2019, gorsline2021}. Besides reducing the chances of data stealing, this also increases the users' confidence in applications and services as it lessens the need to provide personal data to a third party. Furthermore, edge computing enables ML systems to be more predictable as the latency is not dependent on the available network bandwidth \cite{stoica2017, bernhard2019, gorsline2021}. Therefore, edge computing enables developers to provide reliable throughput guarantees, which are mandatory in real-time or mission-critical scenarios \cite{papernot2018}.

Edge devices are now the new frontier for ML deployment \cite{stoica2017, bernhard2019, gorsline2021, costa2022_2}. Of the panoply of devices typically deployed at the edge of the network, microcontrollers (MCUs) are getting special attention, as newer architectures aligned with friendly developer tools (e.g., CMSIS-NN \cite{cmsis-nn}, PULP-NN \cite{pulp-nn}, and TensorFlow (TF) Lite Micro \cite{tflitemicro}) are enabling the deployment of ML systems in devices that are several orders of magnitude cheaper and less power-hungry than application processors (APUs) or graphical processing units (GPUs). Despite the resource-constrained nature of these devices, artificial neural network (ANN) compression techniques such as quantization have proved to have negligible impact on accuracy while tremendously improving latency \cite{cmsis-nn, pulp-nn}. 

In parallel with this development, recent research has shown that full-precision ANNs are vulnerable to the so-called adversarial examples \cite{yuan2019, wang2019, zhou2022}, i.e., inputs carefully crafted to fool the neural network into making an obvious misclassification. In this context, it is essential to ask if quantized neural networks (QNNs) are susceptible to this class of attacks and, if so, how QNNs could be protected. Previous research found that QNNs are still vulnerable to gradient-free attacks and adversarial example transferability, even to those crafted on full-precision ANNs using weak attacks such as Fast Gradient Sign Method (FGSM) \cite{bernhard2019, gorsline2021, zhao2018, lin2019, duncan2020}. \textcite{lin2019} pioneered in adversarial defenses for QNNs by proposing a novel training algorithm that reduces the quantization-shift phenomenon; however, the authors do not provide a clear picture of the scope of their work. As GPUs operate better with integer precision, QNNs are deployed not only on ultra-low-power MCUs but also on powerful GPUs. As GPUs have far more memory and computing power than ultra-low-power MCUs, QNNs deployed on these devices show significant differences in the quantization policy, kernel filter size, and network depth. In this context, providing a clear picture of the quantization policy and the target computing devices when working with QNNs is fundamental. To the best of our knowledge, no work proposes possible defense mechanisms or evaluates the feasibility of adapting the existing defenses for full-precision ANNs on QNNs tailored for the TinyML scenario.

To fill this gap, we performed a comprehensive empirical evaluation to understand how attacks and defenses targeting full-precision ANNs perform in the presence of QNNs. Despite the few studies on the adversarial robustness of QNNs, the information tends to be scattered and needs a well-defined basis for ground comparison. Furthermore, if adversarial examples also affect QNNs, as previously stated, it is paramount to paint a clear picture of QNN defenses. This paper tests the robustness of three QNNs targeting TinyML applications on ten attacks and six possible defenses. This empirical evaluation aims to answer the following research questions:
\begin{itemize}
    \item \textit{Q1 - Can quantization alone protect neural networks from adversarial attacks?} 
    \item \textit{Q2 - How does the bit-width of quantization affect a QNN's defense against adversarial examples?}
    \item \textit{Q3 - Do defenses for full-precision ANNs still work for QNNs?}
    \item \textit{Q4 - Can existing defenses for full-precision ANNs be used on resource-constrained MCUs (deep/extreme edge)?}
\end{itemize}

To answer \textit{Q1} (Section \ref{sec:attacks_eval}), we evaluated the robustness of QNNs without any defense in place. We measured the adversarial accuracy and the distortion when directly attacking QNNs and how adversarial examples crafted on full-precision ANNs transfer to QNNs. We observe that quantization increases the average point distance to the decision boundary and makes it more difficult for attacks to optimize over the loss surface, leading to the explosion or vanishing of the estimated gradient. Additionally, we observe poor transferability of adversarial examples crafted on full-precision ANNs to QNNs, resulting from gradient misalignment and quantization-shift. Overall, quantization minimizes the effects of small perturbations but can amplify bigger perturbations.

To answer \textit{Q2} (Section \ref{sec:attacks_eval}), we replicated all selected attacks on 8-bit and 16-bit QNNs. We evaluated how the bit-width affects the adversarial example transferability and the success of direct attacks. We observe that the average point distance to the decision boundary increases as the neural network precision decreases. This explains why int-8 models are generally more robust than int-16 and float-32 models, with int-16 being more robust than float-32. Furthermore, int-16 models are more affected than int-8 models by the adversarial examples transferred from float-32 ANNs. This occurs as int-8 QNNs map more float-32 values in the same quantization bucket than int-16 QNNs, which enhances the effect of quantization-shift and gradient misalignment.

To answer \textit{Q3} and \textit{Q4} (Section \ref{sec:defenses_eval}), we tested six popular defense mechanisms against the attacks that impact the most the adversarial accuracy of QNNs before any defense. We observe that input pre-processing defenses show impressive results in denoising small perturbations but fall short when the perturbation increases. Train-based defenses generally increase the average point distance to the decision boundary, even after quantization. However, not all defenses can be fully applied in ultra-low-power ML applications. Train-based defenses are easily portable, as they do not require additional computing power at inference. However, they are less robust than pre-processing defenses for smaller, fine-grained perturbations. 

In short, this work makes the following contributions:

\begin{itemize}
    \item \textit{Presents the most comprehensive empirical evaluation of adversarial robustness of QNNs targeting TinyML applications (Section \ref{sec:attacks_eval});} 
    \item \textit{Evaluates the impact of bit-width on adversarial example robustness, testing 8-bit and 16-bit quantization setting of TF Lite Micro (Section \ref{sec:attacks_eval});}
    \item \textit{Delves into the uncharted territory of defenses against adversarial examples in the context of QNNs and empirically assesses if defenses designed for full-precision networks still work when applied to QNNs (Section \ref{sec:defenses_eval})}.
\end{itemize}

\section{Motivation and Methodology}
\subsection{TinyML: The New Wave in AI} \label{sec:motivation}
TinyML is a field of ML dedicated to the design and development of hardware and software solutions to enable the execution of ML inference under a milliWatt scale \cite{tinyml}. By enabling on-device and near-sensor data analysis, TinyML eliminates the imperative reliance on the cloud infrastructure for inference results \cite{bernhard2019, gorsline2021}. This paradigm shift not only enables the dissemination of ML in regions without reliable network, but also ensures greater responsiveness and bolstered privacy.

Under this umbrella, ML at the extreme edge (powered by MCUs) is one of the hottest topics in the ML field due to the proliferation of these devices in our infrastructures/society. In effect, ARM predicts that MCUs are expected to reach 1 trillion across various market segments by 2035 \cite{cmsis-nn}. A recent report \cite{abi2022} from ABI Research dating to 2022 emphasized that TinyML Software-as-a-Service was expected to exceed US\$220 million in revenue in 2022 and have the potential to become a billion-dollar market by 2030. This same company also forecasts that the TinyML market will grow from 15.2 million shipments in 2020 to 2.5 billion in 2030 \cite{abi2021}.

In academia, this trend has fueled the development of multiple software and hardware solutions to enable and optimize the execution of TinyML and has fuelled the rise of RISC-V instruction set architecture (ISA). Among other relevant works, \textcite{gap8} proposed a custom SoC based on the emerging RISC-V ISA \cite{gap8}, while \textcite{pulp-nn} proposed a software library that optimizes neural network functions on this same SoC. \textcite{scherer22} proposed a RISC-V hardware accelerator able to deliver 3200 inferences/second while only consuming 12.2 milliWatts.

In terms of use cases, TinyML has been widely deployed for (i) word spotting, (ii) object recognition, (iii) object counting, and (iv) audio/voice detection \cite{tinyml, abi2022}. More recently, its application has been extended to the detection of (i) forest fire, (ii) shape, and (iii) seizure \cite{abi2022}.

\subsection{Scope} \label{sec:scope}
This work targets ultra-low-power ML applications (TinyML), commonly powered by Arm Cortex-M (Armv7-M or Armv8-M) and RISC-V MCUs (e.g., PULP \cite{pulp}). These devices typically have a single core and feature Flash and static random access memories (SRAM). They execute instructions in place from physical memory and lack a memory management unit (MMU) for virtual memory support. Consequently, they are unable to run Unix-like operating systems. 

TF Lite Micro, which operates on top of the ARM-developed CMSIS-NN library, represents the state-of-the-art solution for executing QNNs on these MCUs. TF Lite Micro is accelerated by ARM’s single instruction multiple data (SIMD); however, it requires a quantization policy that has been mostly neglected in previous works. We highlight that quantization can be used (i) to deploy QNNs in MCUs, (ii) to speed up inference on GPUs or tensor processing units (TPUs), or (iii) to simply reduce the size of neural networks in mobile devices (e.g., smartphones). However, the quantization policy for GPU/TPUs and application processors (APUs) is significantly different from MCUs - thus, QNNs have completely different architectures and parameters, leading to different findings.

To help standardize the throughput of ultra-low-power hardware on ML applications, \textcite{tinyml} developed the MLPerf Tiny benchmark, composed of four QNNs targeting image and sound applications. Our empirical evaluation targets QNNs developed under the exact specification as MLPerf Tiny: QNNs using fixed-point quantization with 8 or 16 bits for weights and activations and quantized according to the TF Lite Micro policy. Within this policy, the number of bits representing the integer and fractional parts is not fixed, varying for weights and activations within the same layer.

\subsection{Adversary Model}
We define the adversary model of our research along three dimensions, which concern the (i) goal, (ii) capability, and (iii) knowledge of the attacker.

\mypara{Adversary Goal.} The adversary's goal is to compromise the decision integrity of a supervised model, disrupting severe accuracy drops. Previous research shows that the attacker can achieve this goal by interfering with the training or inference phase. However, research on training at the deep edge is still at an embryonary stage \cite{costa2022}, and this class of devices is still only used for inference purposes. The most relevant part of the training of the models deployed on deep-edge devices is still performed in cloud-like environments. Therefore, we limit the scope of our research to the inference phase. Attacks to craft adversarial examples can either be targeted or non-targeted. Targeted attacks misguide an ANN by making it classify an adversarial sample as a specific class. In contrast, in non-targeted attacks, the class returned by the model is arbitrary except for the original one. Our research encompasses only non-targeted attacks, as they enable a broader assessment of the robustness and generalization of a neural network.

\mypara{Adversary Capability.} Capability refers to the attacker's capacity to control or manipulate the input data. At inference, the generic data processing pipeline of an ML system considers that the (i) input features are collected from sensors or data repositories and (ii) processed in the digital domain, (iii) used by the model to infer about the surrounding environment, (iv) which communicates the result to the appropriate actuator \cite{papernot2018}. Considering this pipeline, adversarial examples can be generated either in the physical domain, by tampering with sensors, or in the digital domain. This work only considers attacks and defenses in the digital domain, which dominates the available literature.

\mypara{Adversary Knowledge.} Datasets are, at least, divided into two parts: one for training and another for testing. The training subset is exclusively used for training and the testing subset is used to evaluate the real accuracy of the model. In our threat model, we assume that the attacker has full access to the testing data and can use this data as input to a given attack to craft an adversarial dataset, on which the accuracy of the attacked model is evaluated. Regarding the knowledge of the target model, we define three testing scenarios. For every testing scenario, the adversary is aware of the quantization status of the network, i.e., the adversary knows that the neural network under attack is quantized.

\myparait{White-box:} The adversary has full access to the attacked model, including its architecture and parameters. When considering QNNs, the adversary also has access to the scaling and zero-point factors used to quantize the input, output, and internal parameters.

\myparait{Gray-box:} The adversary does not know the architecture of the target model or its internal parameters. Nevertheless, they have access to class probabilities. The scaling and zero-point factors used in quantization remain secret.

\myparait{Black-box:} The adversary can only feed input data to the target model and query its output. They cannot access the class probability vector but the final decision.

\subsection{Selected Datasets and Models}
To benchmark the throughput of the hardware used in TinyML, \textcite{tinyml} proposed the MLPerf Tiny benchmark. This benchmark results from the collaborative effort of more than 30 organizations from industry and academia and reflects the community's needs by simulating real-world applications (see Section \ref{sec:motivation}). As the datasets and models used in these applications are the companies' intellectual property, researchers selected representative and open-sourced models and datasets. 

We borrowed the models and datasets from the MLPerf Tiny benchmark but limited the empirical evaluation to image processing tasks, as images provide a convenient means to observe the impact of adversarial attacks and defenses. Table \ref{tab:bench_suite} summarizes the benchmarking suite.

\begin{table}[h]
\centering
\caption{Benchmarking suite}
\label{tab:bench_suite}
\resizebox{\columnwidth}{!}{%
\begin{tabular}{@{}lll@{}}
\toprule
\multicolumn{1}{l}{\textbf{Use Case}} & \multicolumn{1}{l}{\textbf{Dataset}} & \multicolumn{1}{l}{\textbf{Model}} \\ \midrule
Image classification                  & CIFAR-10                             & ResNet-8                           \\
Human face detection                     & Visual Wake Words (VWW)            & DS-CNN                             \\
Coffee plant monitoring               & Coffee Dataset                       & DS-CNN                             \\ \bottomrule
\end{tabular}
}
\end{table}

In terms of image processing, MLPerf Tiny comprises two datasets: (i) CIFAR-10 and (ii) Visual Wake Words (VWW). CIFAR-10 consists of 60000 32x32 red-green-blue (RGB) images distributed over ten classes, each representing a different object. There are 50000 images for training and 10000 images for testing. VWW is a binary dataset with 115000 96x96 RGB images representing a vision use-case of identifying whether a person is present in the image or not. The training subset contains 98568 images, and the testing subset contains 10961.

We included a third dataset because evaluating attacks and defenses on only two datasets may fall short. For this purpose, we borrowed the work \cite{lucan2022}, which lists and reviews ML implementations for Arm Cortex-M MCUs. After analyzing the reproducibility of each listed work, we replicated the work \cite{vita2020}. This work addressed the development of an ML system devoted to detecting coffee plant diseases. The dataset is public and contains 1747 images for training and testing distributed over four classes, each corresponding to a different disease. All QNNs included in our study are quantized post-training.

\subsection{Selected Attacks and Defenses}
We prioritize attacks/defenses with great impact and citation in the research community and whose implementation is open-sourced. Furthermore, since we target QNNs following the TF Lite Micro quantization policy, we prioritize attacks and defenses implemented in TensorFlow 2 (TF2) or Keras (the backend engine of TF2). TF2 provides straightforward mechanisms to operate with models in TF Lite format. We also prioritize attacks/defenses available in ART \cite{art}, Cleverhans \cite{cleverhans}, or Foolbox \cite{foolbox1, foolbox2} frameworks, as these frameworks provide a straightforward mechanism to benchmark different attacks and defenses upon a common interface. Our evaluation suite includes six white-box, two gray-box, and two black-box attacks, described in Table \ref{tab:state_of_art_attacks}.

\mypara{White-box Attacks.} White-box attacks are used to evaluate the robustness of QNNs against adversarial transferability (Section \ref{sec:res_transferability}). Transferability is a well-known vulnerability of ANNs and refers to the ability of an adversarial example to simultaneously evade the prediction on two or more ML models performing similar tasks (trained on similar or equal datasets). In our threat model, the adversary uses the white-box attacks listed in Table \ref{tab:state_of_art_attacks} to craft adversarial examples on a full-precision ANN, which are then quantized and applied against the corresponding int-16 and int-8 QNNs. The full-precision ANN can be seen as a surrogate model mimicking the QNN on which the adversary performs transfer-based attacks.

We do not use white-box attacks to directly attack QNNs as white-box attacks require an exact gradient calculation, which is impossible due to the quantization policy. More specifically, this results from the dynamic bit-width used to represent the fractional and integer parts throughout the network. Despite every activation and model parameter using a static bit-width of 8 or 16 bits, the number of bits dedicated to the fractional and integer parts is dynamic. Furthermore, the fractional bit-width in the same layer differs for the input, output, and model parameters. In this scenario, the only feasible way to calculate a gradient is to convert these parameters to a common format, preferably with higher precision (float-32), and calculate the gradient afterward. This process corresponds to dequantizing the QNN back to full-precision and applying the attacks afterward. The resulting adversarial examples are then quantized (converted from float-32 to int-16 or int-8) and transferred to the respective QNN. However, this process is, in reality, evaluating the transferability of adversarial examples.

\mypara{Gray- and Black-Box Attacks.} The gray- and black-box attacks listed in Table \ref{tab:state_of_art_attacks} are used to evaluate the adversarial transferability, but also to directly attack each ANN and QNN, i.e., craft adversarial examples tailored for a given neural network, and compare how the bit-width of a neural network affects its adversarial robustness. We do not include white-box attacks targeting QNNs as existing attacks only apply to QNNs whose gradients can be approximated by Straight-Through Estimator (STE) or Mirror Descent techniques. This feature is absent in QNNs compliant with the TF Lite Micro quantization policy. Nevertheless, we include a series of gray- and black-box attacks (Square Attack, Boundary Attack, and GeoDA) developed to tackle gradient obfuscation, a phenomenon disrupted by QNNs that gives a false sense of security.

\mypara{Defenses.} We consider six defense mechanisms (Table \ref{tab:state_of_art_defenses}) distributed across the two types of defense proposed by the research community. Our empirical evaluation includes four train-based defenses and the other two based on input pre-processing. Although there might be more recent defenses available in the literature than those we selected, PGD Adversarial Training, proposed in 2018 and included in our work, remains the state-of-the-art defense against adversarial attacks according to recent studies \cite{autoattack, zhang2021_2, ferrari23}. Nevertheless, we still include a recent iteration of this defense, entitled Sinkhorn Adversarial Training (2021), which promises to reduce the ground distance between original images and their adversarial representations.

\newbox\jsmaEquation
\sbox\jsmaEquation{\( S=\begin{Bmatrix}
0 \text{, if } \frac{\partial L_{y'}(x)}{\partial x_i} < 0 \text{ or } \sum_{y\neq y'}^{} \frac{\partial L_y(x)}{\partial x_i} > 0 \\ \frac{\partial L_{y'}(x)}{\partial x_i} * \left| \sum_{y \neq y'}^{} \frac{\partial L_y(x)}{\partial x_i} \right| \text{, otherwise}
\end{Bmatrix} \)}

\begin{table*}[!t]
\caption{Adversarial example attacks}
\label{tab:state_of_art_attacks}
\footnotesize
\resizebox{\textwidth}{!}{%
\begin{tabular}{llllll}
\hline
\multicolumn{1}{l|}{\textbf{Attack}} &
  \multicolumn{1}{l|}{\textbf{Year}} &
  \multicolumn{1}{l|}{\textbf{Norm}} &
  \multicolumn{1}{l|}{\textbf{Description}} &
  \textbf{Relevant Formulas} \\ \hline
\rowcolor[HTML]{C0C0C0} 
\multicolumn{5}{c}{\cellcolor[HTML]{C0C0C0}\textit{\textbf{WHITE-BOX}}} \\ \hline

\multicolumn{1}{l|}{\textbf{DeepFool~\cite{deepfool}}} &
  \multicolumn{1}{l|}{2016} &
  \multicolumn{1}{l|}{$L_2$} &
  \multicolumn{1}{l|}{\begin{tabular}[c]{p{0.4\textwidth}}DeepFool is an iterative attack that seeks to find the minimum perturbation $\delta$ to misclassify $x$, moving it toward the closest boundary. At iteration $t$, DeepFool uses Taylor expansion to linearize the boundary around $x_t$, moving $x_t$ outward. This process is repeated until $x_t$ is misclassified.\end{tabular}} &
  $\displaystyle min||\delta_t|| \text{ such that } L(x'_t) + \bigtriangledown L(x'_t)^T \delta_t = 0$ \\

\rowcolor[HTML]{EFEFEF} 
\multicolumn{1}{l|}{\textbf{\begin{tabular}[c]{@{}l@{}}JSMA~\cite{jsma}\end{tabular}}} &
  \multicolumn{1}{l|}{2016} &
  \multicolumn{1}{l|}{$L_0$} &
  \multicolumn{1}{l|}{\begin{tabular}[c]{p{0.4\textwidth}}The Jacobian-based Saliency Map Approach (JSMA) iteratively perturbs the most relevant features of $x$. It first calculates the Jacobian matrix of $x$ and then the adversarial saliency map $S$. JSMA distorts features in order of relevance: each feature is distorted by $\delta$ until $x$ is misclassified or the maximum number of features is distorted.\end{tabular}} &
\usebox\jsmaEquation \\

\multicolumn{1}{l|}{\textbf{\begin{tabular}[c]{@{}l@{}}C\&W~\cite{cw}\end{tabular}}} &
  \multicolumn{1}{l|}{2017} &
  \multicolumn{1}{l|}{\begin{tabular}[c]{@{}l@{}}$L_0$\\ $L_2$\\ $L_\infty$\end{tabular}} &
  \multicolumn{1}{l|}{\begin{tabular}[c]{p{0.4\textwidth}}For any $L_p$ norm, the Carlini and Wagner (C\&W) attack follows the same optimization problem to find the minimum distortion ($min_\delta$). $c$ is a constant weighting the relative importance of distance and loss, and $Z(x')_y$ denotes the probability of the target class $y$. $k$ sets the minimum confidence for the adversarial example.\end{tabular}} &
  \begin{tabular}[c]{@{}l@{}} $\displaystyle min_\delta \, D(x, x+\delta) + c . L(x+\delta)$ \\ \\ $L(x') = max(max\{Z(x')_y: y \neq y'\} - Z(x')_y -k)$ \\ \\ $\delta_i = \frac{1}{2}(tanh(\sigma_i)+1)-x_i$ \end{tabular}  \\

\rowcolor[HTML]{EFEFEF} 
\multicolumn{1}{l|}{\textbf{\begin{tabular}[c]{@{}l@{}}PGD~\cite{pgd}\end{tabular}}} &
  \multicolumn{1}{l|}{2018} &
  \multicolumn{1}{l|}{$L_\infty$} &
  \multicolumn{1}{l|}{\begin{tabular}[c]{p{0.4\textwidth}}Projected Gradient Descent (PGD) is an iterative attack that applies a small perturbation $\alpha$ to each feature of $x$ in the same direction as the gradient till a maximum distortion $\delta$. At the beginning of each iteration, PGD adds random noise to $x$ and restarts the iteration from the most adversarial point.\end{tabular}} &
  $\displaystyle x_t = Clip_{x_t,\delta }(x_{t-1} + \alpha * Sign\frac{\partial L(w,x_{t-1},y)}{\partial x_{t-1}})$  \\
  
\multicolumn{1}{l|}{\textbf{\begin{tabular}[c]{@{}l@{}}EAD~\cite{ead}\end{tabular}}} &
  \multicolumn{1}{l|}{2018} &
  \multicolumn{1}{l|}{\begin{tabular}[c]{@{}l@{}}$L_1$\\ $L_2$\end{tabular}} &
  \multicolumn{1}{l|}{\begin{tabular}[c]{p{0.4\textwidth}}Elastic-Net Attack to DNNs (EAD) is a variant of the C\&W attack that simultaneously minimizes $L_1$ and $L_2$ norms. $c$ weights the loss function $L$ against the distortion and $\beta$ sets the relevance of $L_1$ norm.\end{tabular}} &
  $\displaystyle min_{x'} c * L(x',y') + \beta\left \| x'-x \right \|_1 + \left \| x'-x \right \|_2^2$ \\ 


\rowcolor[HTML]{EFEFEF} 
\multicolumn{1}{l|}{\cellcolor[HTML]{EFEFEF}\textbf{\begin{tabular}[c]{@{}l@{}}AutoAttack~\cite{autoattack}\end{tabular}}} &
  \multicolumn{1}{l|}{\cellcolor[HTML]{EFEFEF}2020} &
  \multicolumn{1}{l|}{\begin{tabular}[c]{@{}l@{}}$L_1$\\ $L_2$\\ $L_\infty$\end{tabular}} &
  \multicolumn{1}{l|}{\cellcolor[HTML]{EFEFEF}\begin{tabular}[c]{p{0.4\textwidth}}AutoAttack is a parameter-free attack, grouping four attacks: two Auto-PGD (APGD) versions, Fast Adaptive Boundary Attack (FAB), and SA. APGD differs from PGD as the step size is adjusted dynamically based on the optimization trend. If the loss increases in 75\% or more of the past iterations, the step size is maintained; otherwise, it is halved. The two versions of APGD differ in their loss function: one uses cross-entropy, and the other uses the difference of logits ratio. FAB is an optimized version of DeepFool.\end{tabular}} &
  -  \\ \hline

\rowcolor[HTML]{C0C0C0} 
\multicolumn{5}{c}{\cellcolor[HTML]{C0C0C0}\textit{\textbf{GRAY-BOX}}} \\ \hline
\multicolumn{1}{l|}{\textbf{\begin{tabular}[c]{@{}l@{}}ZOO~\cite{zoo}\end{tabular}}} &
  \multicolumn{1}{l|}{2017} &
  \multicolumn{1}{l|}{$L_2$} &
  \multicolumn{1}{l|}{\begin{tabular}[c]{p{0.4\textwidth}}Zeroth-Order Optimization (ZOO) is a gray-box version of C\&W$_{L2}$. ZOO uses an alternative formula to estimate the gradient, where $h$ is a small constant and $e_i$ is a standard basis vector with only the $i^{th}$ component set to 1. To be efficient, ZOO uses stochastic coordinate descent methods.\end{tabular}} &
  $\displaystyle \frac{\partial f(x)}{\partial x_i} \approx \frac{f(x+he_i) - f(x-he_i)}{2h}$ \\

\rowcolor[HTML]{EFEFEF} 
\multicolumn{1}{l|}{\textbf{\begin{tabular}[c]{@{}l@{}}SA~\cite{sa}\end{tabular}}} &
  \multicolumn{1}{l|}{2020} &
  \multicolumn{1}{l|}{\begin{tabular}[c]{@{}l@{}}$L_2$\\ $L_\infty$\end{tabular}} &
  \multicolumn{1}{l|}{\begin{tabular}[c]{p{0.4\textwidth}}Square Attack (SA) is an iterative attack that samples a random square-shaped perturbation per iteration and adds it to $x$ if it increases the classifier's error. The attack stops as soon as it finds an adversarial example. SA decreases the width and distortion of squares according to a fixed schedule.\end{tabular}} &
  - \\ \hline
  
\rowcolor[HTML]{C0C0C0} 
\multicolumn{5}{c}{\cellcolor[HTML]{C0C0C0}\textit{\textbf{BLACK-BOX}}} \\ \hline
\multicolumn{1}{l|}{\textbf{\begin{tabular}[c]{@{}l@{}}BA~\cite{ba}\end{tabular}}} &
  \multicolumn{1}{l|}{2018} &
  \multicolumn{1}{l|}{$L_2$} &
  \multicolumn{1}{l|}{\begin{tabular}[c]{p{0.4\textwidth}}Boundary Attack (BA) samples a first adversarial example $x'$ from a uniform distribution. Then it performs a random walk along the decision boundary such that $x'$ remains adversarial and the distance towards $x$ decreases. BA samples a new direction at each iteration and projects $x'_i$ on the $\delta$ neighbor-ball of $x$. Then $x'_i$ is moved towards $x$. The $x'_i$ closest to $x$ respecting the adversarial condition is selected.\end{tabular}} & - \\
  
\rowcolor[HTML]{EFEFEF} 
\multicolumn{1}{l|}{\cellcolor[HTML]{EFEFEF}\textbf{\begin{tabular}[c]{@{}l@{}}GeoDA~\cite{geoda}\end{tabular}}} &
  \multicolumn{1}{l|}{\cellcolor[HTML]{EFEFEF}2020} &
  \multicolumn{1}{l|}{\cellcolor[HTML]{EFEFEF}$L_2$} &
  \multicolumn{1}{l|}{\cellcolor[HTML]{EFEFEF}\begin{tabular}[c]{p{0.4\textwidth}}Geometric Decision-Based Attack (GeoDA) uses binary search to find a point belonging to the nearest decision boundary. This point is used to define a hyperplane that approximates the boundary surface. After defining the hyperplane, GeoDA uses an iterative method to find a vector normal to the boundary and uses that vector to perturb $x$ until misclassification.\end{tabular}} & - \\ \hline
\end{tabular}%
}
\end{table*}
\begin{table*}[]
\centering
\caption{Adversarial Example Defenses. Basic Iterative Method (BIM) is an iterative version of FGSM.}
\label{tab:state_of_art_defenses}
\resizebox{\textwidth}{!}{%
\begin{tabular}{llll}
\hline
\multicolumn{1}{l|}{\textbf{Defense}} &
  \multicolumn{1}{l|}{\textbf{Year}} &
  \multicolumn{1}{l|}{\textbf{Description}} &
  \textbf{Tested Against} \\ \hline
\rowcolor[HTML]{C0C0C0} 
\multicolumn{4}{c}{\cellcolor[HTML]{C0C0C0}\textit{\textbf{TRAIN-BASED DEFENSES}}} \\ \hline
\multicolumn{1}{l|}{\textbf{\begin{tabular}[c]{@{}l@{}}Defensive\\ Distillation\\ \cite{defensive_distil}\end{tabular}}} &
  \multicolumn{1}{l|}{2016} &
  \multicolumn{1}{l|}{\begin{tabular}[c]{p{0.7\textwidth}}Distillation works by firstly training the teacher model with the original dataset using the hard-coded labels. Then, the soft labels returned by the teacher model are used to train the student model. Defensive Distillation uses high distillation temperatures in softmax activation, as high temperatures cause the softmax to return smoother probabilities, reducing the sensitivity of the ANN to input variations.\end{tabular}} &
  JSMA \\
  
\rowcolor[HTML]{EFEFEF} 
\multicolumn{1}{l|}{\cellcolor[HTML]{EFEFEF}\textbf{\begin{tabular}[c]{@{}l@{}}PGD\\ Adversarial\\ Training\\ \cite{pgd}\end{tabular}}} &
  \multicolumn{1}{l|}{\cellcolor[HTML]{EFEFEF}2018} &
  \multicolumn{1}{l|}{\begin{tabular}[c]{p{0.7\textwidth}}Adversarial training consists in retraining an ANN by injecting adversarial examples into the training dataset. The first work on adversarial training \cite{fgsm} proposed to generate adversarial examples using FGSM in each training iteration, mixing clean samples and adversarial examples. This work was later borrowed by \cite{pgd}, which proposed adversarial training with PGD.\end{tabular}} &
  \begin{tabular}[c]{@{}l@{}}FGSM\\ BIM\\ DeepFool\\ JSMA\\ C\&W Attack\end{tabular} \\
  
\multicolumn{1}{l|}{\textbf{\begin{tabular}[c]{@{}l@{}}Ensemble\\ Adversarial\\ Training\\ \cite{ens_adv_train}\end{tabular}}} &
  \multicolumn{1}{l|}{2018} &
  \multicolumn{1}{l|}{\begin{tabular}[c]{p{0.7\textwidth}}Ensemble Adversarial Training improves the training dataset by incorporating adversarial examples from other classifiers. This approach separates the generation of adversarial examples from the training process, enabling it to scale more efficiently to larger datasets than PGD Adversarial Training.\end{tabular}} &
  \begin{tabular}[c]{@{}l@{}}FGSM\\ BIM\\ PGD\end{tabular} \\

\rowcolor[HTML]{EFEFEF} 
\multicolumn{1}{l|}{\cellcolor[HTML]{EFEFEF}\textbf{\begin{tabular}[c]{@{}l@{}}Sinkhorn\\ Adversarial\\ Training\\ \cite{sat}\end{tabular}}} &
  \multicolumn{1}{l|}{\cellcolor[HTML]{EFEFEF}2021} &
  \multicolumn{1}{l|}{\begin{tabular}[c]{p{0.7\textwidth}}Sinkhorn Adversarial Training (SAT) is a recent iteration of PGD Adversarial Training that adopts Sinkhorn Divergence as the loss function. The Sinkhorn Divergence exploits the optimal transport theory to calculate and reduce the ground distance between representations of original images and their adversarial counterparts. Adversarial images are calculated at each training epoch using the PGD attack.\end{tabular}} &
  \begin{tabular}[c]{@{}l@{}}PGD\end{tabular} \\ \hline

\rowcolor[HTML]{C0C0C0} 
\multicolumn{4}{c}{\cellcolor[HTML]{C0C0C0}\textit{\textbf{INPUT PRE-PROCESSING DEFENSES}}} \\ \hline
\multicolumn{1}{l|}{\textbf{\begin{tabular}[c]{@{}l@{}}Feature\\ Squeezing\\ \cite{feat_squeezing}\end{tabular}}} &
  \multicolumn{1}{l|}{2017} &
  \multicolumn{1}{l|}{\begin{tabular}[c]{p{0.7\textwidth}}Feature Squeezing decreases the color bit-depth and performs spatial smoothing of the input sample. If the classification of squeezed and unsqueezed samples differ by a large margin, i.e., exceeds a given $L_1$ distance, Feature Squeezing classifies the sample as adversarial.\end{tabular}} &
  Transferability \\
  
\rowcolor[HTML]{EFEFEF} 
\multicolumn{1}{l|}{\cellcolor[HTML]{EFEFEF}\textbf{\begin{tabular}[c]{@{}l@{}}Pixel\\ Defend\\ \cite{pixel_defend}\end{tabular}}} &
  \multicolumn{1}{l|}{\cellcolor[HTML]{EFEFEF}2018} &
  \multicolumn{1}{l|}{\begin{tabular}[c]{p{0.7\textwidth}}Pixel Defend uses a generative network before the classifier to denoise adversarial examples. The generative network is trained on the same data as the classifier to learn the inherent data distribution. This model is then used to reconstruct adversarial examples, bringing them back to the same distribution of training data. Pixel Defend uses PixelCNN as the generative network.\end{tabular}} &
  \begin{tabular}[c]{@{}l@{}}FGSM\\ BIM\\ DeepFool\\ C\&W Attack\end{tabular} \\ \hline
\end{tabular}%
}
\end{table*}

\subsection{Empirical Evaluation}
To answer the research question \emph{Q1}, we evaluate how QNNs, without any defense in place, perform compared to the respective full-precision ANNs in the face of adversarial examples. The evaluation focuses on adversarial accuracy, i.e., the percentage of adversarial examples correctly classified, and on adversarial distortion. To measure the distortion of adversarial examples, we rely on the distance metrics $L_0$, $L_1$, $L_2$, and $L_\infty$. We compare the robustness of QNNs and ANNs when facing adversarial examples specifically tailored for the neural network under attack (Section 3.1) and how adversarial examples crafted on full-precision ANNs transfer to QNNs (Section 3.2). To answer the research question \emph{Q2}, the experiments are performed for QNNs with 8-bits and QNNs with 16-bits. Our work targets the quantization policy of TF Lite Micro, which restricts the bit-width to these two options.

Table \ref{tab:attack_config} (Appendix \ref{ap:attack_defense_config}) details the configuration of the attacks. We configured the attacks such that the accuracy of the target ANN drops below the value of random choice (10\% for CIFAR-10, 50\% for VWW, and 25\% for Coffee). Then, we applied the attacks with the same parameters to the respective QNN (direct attacks) or quantized the adversarial examples previously produced and applied them to QNNs (adversarial transferability). Other possible approaches to configure the attacks would be to (i) stop the attack when the distortion reaches an upper bound instead of stopping the attack upon a fixed accuracy degradation or (ii) stop the attack only when the adversarial accuracy drops to 0\% in every model/bit-width combo and then compare the distortion. Regarding the former, it must be noted that not all attacks optimize the same distortion ($L_p$) metric. Considering that there are 4 distortion metrics, this approach will increase the engineering effort of our work as we would have to control 4 variables ($L_0$, $L_1$, $L_2$, and $L_\infty$) instead of 1 (accuracy). Regarding the latter, it would require distinct configurations of the attacks for different bit-widths, which would require more engineering effort.

To answer the research questions \emph{Q3} and \emph{Q4}, we split the evaluation of defenses into three parts. Firstly, we apply each defense in full-precision and quantized models. For train-based defenses, we first retrain the full-precision ANN with the defense constraints and then we quantize the resulting ANN, obtaining the retrained QNN. For input pre-processing defenses, we simply apply the defense before the ANN/QNN input. The configuration of each defense is detailed in Table \ref{tab:defenses_config} (Appendix \ref{ap:attack_defense_config}). For each defense, we experimented with a different set of hyper-parameters around the default values suggested by the defenses' authors and selected the configuration that perform the best against the PGD attack. Secondly, we quantitatively test each defense against the attacks that most impact the adversarial accuracy of QNNs. The evaluation includes six attacks (three white-box and three gray- and black-box). Thirdly, we perform a qualitative analysis of each defense. We evaluate whether they are suitable or not for TinyML and if so, which adjustments must be made to fit in the TinyML scenario. The results are detailed in Section 4.

\subsection{Threats to Validity}
The experiments were conducted by two independent researchers on two separate but equal platforms: STM NUCLEO-L552ZE-Q. Whenever possible, we used the implementation of attacks and defenses available on the ART framework \cite{art}, except for AutoAttack and Sinkhorn Adversarial Training. Regarding AutoAttack, the original implementation is significantly different from the implementation of ART. Sinkhorn Adversarial Training is not available on ART and is originally implemented in PyTorch. In these two cases, we borrowed the original implementations and adapted them to work with TF2. Despite all efforts, there may be slight variations compared to the original works. To encourage further reproducibility of our results, all artifacts are available open-source\footnote{\url{https://gitlab.com/ESRGv3/QNN-Adversarial-Robustness.git}}.

\subsection{Summary of Observations}
First, quantization increases the average point distance to the decision boundary and makes it more difficult for attacks to optimize over the loss surface. This effect can lead to either the explosion or vanishing of the estimated gradient, a phenomenon known as gradient masking. Second, adversarial examples crafted on full-precision ANNs do not transfer well to QNNs due to gradient misalignment and quantization-shift. Lastly, quantization can mitigate small perturbations but also amplify bigger perturbations.

While input pre-processing defenses show impressive denoising results for small perturbations, their effectiveness diminishes as the perturbation increases. On the other hand, train-based defenses generally increase the average point distance to the decision boundary and this property remains even after quantization. However, not all defenses can be applied in TinyML applications. Train-based defenses are the most easily portable but less robust for smaller, fine-grained perturbations. The presence of a digital signal processing (DSP) or floating-point unit (FPU) on the target MCU potentiates the implementation of input pre-processing defenses.

\section{Robustness Against Adversarial Examples} \label{sec:attacks_eval}
In this section, we empirically evaluate the robustness of QNNs and compare their performance against their full-precision ANNs. Table \ref{tab:attack_config} (Appendix \ref{ap:attack_defense_config}) details the configuration of the attacks. We compare the effect of direct attacks and the adversarial example transferability from float-32 ANNs to int-16 and int-8 QNNs. For both scenarios, we evaluate the adversarial accuracy and distortion under $L_0$, $L_1$, $L_2$, and $L_\infty$ norms. Table \ref{tab:models_test_acc} details the accuracy of the models under test before any attack.

\begin{table}[h]
\footnotesize
\centering
\caption{Test accuracy of the models before attack}
\label{tab:models_test_acc}
\begin{tabular}{@{}lccc@{}}
\toprule
 & \multicolumn{1}{l}{\textbf{CIFAR-10}} & \multicolumn{1}{l}{\textbf{VWW}} & \multicolumn{1}{l}{\textbf{Coffee}} \\ \midrule
\textbf{Float-32} & 88.30\% & 91.30\% & 97.00\% \\
\textbf{Int-16}   & 88.20\% & 91.00\% & 97.00\% \\
\textbf{Int-8}    & 87.50\% & 91.00\% & 97.00\% \\ \bottomrule
\end{tabular}
\end{table}
\begin{table*}[]
\caption{Quantization impact on direct attacks. Red highlights samples that affect the accuracy and whose noise is detectable to the human eye. Yellow/blue highlights samples that do not affect the accuracy and whose noise is detectable/not detectable to the human eye.}
\label{tab:direct_attacks}
\resizebox{\textwidth}{!}{%
\begin{tabular}{lllllllllllllllll}
\hline
\textbf{Attack} &
  \multicolumn{1}{l|}{} &
  \multicolumn{5}{c|}{\textbf{CIFAR-10}} &
  \multicolumn{5}{c|}{\textbf{Visual Wake Words}} &
  \multicolumn{5}{c}{\textbf{Coffee Dataset}} \\ \cline{3-17} 
 &
  \multicolumn{1}{l|}{} &
  \multicolumn{1}{c|}{\textit{Acc}} &
  \multicolumn{1}{c}{$L_0$} &
  \multicolumn{1}{c}{$L_1$} &
  \multicolumn{1}{c}{$L_2$} &
  \multicolumn{1}{c|}{$L_\infty$} &
  \multicolumn{1}{c|}{\textit{Acc}} &
  \multicolumn{1}{c}{$L_0$} &
  \multicolumn{1}{c}{$L_1$} &
  \multicolumn{1}{c}{$L_2$} &
  \multicolumn{1}{c|}{$L_\infty$} &
  \multicolumn{1}{c|}{\textit{Acc}} &
  \multicolumn{1}{c}{$L_0$} &
  \multicolumn{1}{c}{$L_1$} &
  \multicolumn{1}{c}{$L_2$} &
  \multicolumn{1}{c}{$L_\infty$} \\ \hline
\rowcolor[HTML]{C0C0C0} 
\multicolumn{17}{c}{\cellcolor[HTML]{C0C0C0}\textit{\textbf{GRAY-BOX}}} \\ \hline
 &
  \multicolumn{1}{l|}{\textit{Float-32}} &
  \multicolumn{1}{l|}{24.70\%} &
  141.14 &
  4.40 &
  0.31 &
  \multicolumn{1}{l|}{0.05} &
  \multicolumn{1}{l|}{\cellcolor[HTML]{FFCCC9}54.00\%} &
  \cellcolor[HTML]{FFCCC9}13502.90 &
  \cellcolor[HTML]{FFCCC9}462.85 &
  \cellcolor[HTML]{FFCCC9}4.48 &
  \multicolumn{1}{l|}{\cellcolor[HTML]{FFCCC9}0.20} &
  \multicolumn{1}{l|}{\cellcolor[HTML]{FFCCC9}92.00\%} &
  \cellcolor[HTML]{FFCCC9}16291.00 &
  \cellcolor[HTML]{FFCCC9}390.52 &
  \cellcolor[HTML]{FFCCC9}3.02 &
  \cellcolor[HTML]{FFCCC9}0.08 \\
 &
  \multicolumn{1}{l|}{\textit{Int-16}} &
  \multicolumn{1}{l|}{72.00\%} &
  51.97 &
  2.02 &
  0.14 &
  \multicolumn{1}{l|}{0.02} &
  \multicolumn{1}{l|}{\cellcolor[HTML]{FFCCC9}69.00\%} &
  \cellcolor[HTML]{FFCCC9}19269.35 &
  \cellcolor[HTML]{FFCCC9}693.58 &
  \cellcolor[HTML]{FFCCC9}6.60 &
  \multicolumn{1}{l|}{\cellcolor[HTML]{FFCCC9}0.29} &
  \multicolumn{1}{l|}{\cellcolor[HTML]{FFCCC9}80.00\%} &
  \cellcolor[HTML]{FFCCC9}34388.56 &
  \cellcolor[HTML]{FFCCC9}896.46 &
  \cellcolor[HTML]{FFCCC9}6.82 &
  \cellcolor[HTML]{FFCCC9}0.18 \\
\multirow{-3}{*}{\textbf{ZOO}} &
  \multicolumn{1}{l|}{\textit{Int-8}} &
  \multicolumn{1}{l|}{\cellcolor[HTML]{c0d4ef}87.50\%} &
  \cellcolor[HTML]{c0d4ef}0.00 &
  \cellcolor[HTML]{c0d4ef}0.00 &
  \cellcolor[HTML]{c0d4ef}0.00 &
  \multicolumn{1}{l|}{\cellcolor[HTML]{c0d4ef}0.00} &
  \multicolumn{1}{l|}{\cellcolor[HTML]{FFCCC9}84.00\%} &
  \cellcolor[HTML]{FFCCC9}4961.24 &
  \cellcolor[HTML]{FFCCC9}204.20 &
  \cellcolor[HTML]{FFCCC9}1.95 &
  \multicolumn{1}{l|}{\cellcolor[HTML]{FFCCC9}0.09} &
  \multicolumn{1}{l|}{\cellcolor[HTML]{fffc9e}97.00\%} & 
  \cellcolor[HTML]{fffc9e}5168.19 &
  \cellcolor[HTML]{fffc9e}151.97 &
  \cellcolor[HTML]{fffc9e}1.18 &
  \cellcolor[HTML]{fffc9e}0.03 \\
\rowcolor[HTML]{EFEFEF} 
\cellcolor[HTML]{EFEFEF} &
  \multicolumn{1}{l|}{\cellcolor[HTML]{EFEFEF}\textit{Float-32}} &
  \multicolumn{1}{l|}{\cellcolor[HTML]{EFEFEF}47.30\%} &
  3044.04 &
  14.75 &
  0.87 &
  \multicolumn{1}{l|}{\cellcolor[HTML]{EFEFEF}0.22} &
  \multicolumn{1}{l|}{\cellcolor[HTML]{EFEFEF}45.30\%} &
  12721.11 &
  98.78 &
  1.80 &
  \multicolumn{1}{l|}{\cellcolor[HTML]{EFEFEF}0.28} &
  \multicolumn{1}{l|}{\cellcolor[HTML]{EFEFEF}34.10\%} &
  74699.49 &
  218.60 &
  1.80 &
  0.22 \\
\rowcolor[HTML]{EFEFEF} 
\cellcolor[HTML]{EFEFEF} &
  \multicolumn{1}{l|}{\cellcolor[HTML]{EFEFEF}\textit{Int-16}} &
  \multicolumn{1}{l|}{\cellcolor[HTML]{EFEFEF}46.60\%} &
  1139.35 &
  14.01 &
  0.83 &
  \multicolumn{1}{l|}{\cellcolor[HTML]{EFEFEF}0.21} &
  \multicolumn{1}{l|}{42.80\%} &
  12244.77 &
  216.54 &
  2.52 &
  \multicolumn{1}{l|}{0.28} &
  \multicolumn{1}{l|}{35.40\%} &
  64190.22 &
  2715.53 &
  8.57 &
  0.22 \\
\rowcolor[HTML]{EFEFEF} 
\multirow{-3}{*}{\cellcolor[HTML]{EFEFEF}\textbf{Square-$L_2$}} &
  \multicolumn{1}{l|}{\cellcolor[HTML]{EFEFEF}\textit{Int-8}} &
  \multicolumn{1}{l|}{\cellcolor[HTML]{EFEFEF}67.80\%} &
  \cellcolor[HTML]{EFEFEF}443.77 &
  \cellcolor[HTML]{EFEFEF}11.44 &
  \cellcolor[HTML]{EFEFEF}0.75 &
  \multicolumn{1}{l|}{\cellcolor[HTML]{EFEFEF}0.22} &
  \multicolumn{1}{l|}{64.60\%} &
  7176.22 &
  352.30 &
  3.64 &
  \multicolumn{1}{l|}{0.32} &
  \multicolumn{1}{l|}{65.20\%} &
  16830.86 &
  627.71 &
  3.11 &
  0.23 \\
 &
  \multicolumn{1}{l|}{\textit{Float-32}} &
  \multicolumn{1}{l|}{6.10\%} &
  3045.62 &
  33.26 &
  0.65 &
  \multicolumn{1}{l|}{0.01} &
  \multicolumn{1}{l|}{37.60\%} &
  16671.67 &
  248.55 &
  1.78 &
  \multicolumn{1}{l|}{0.01} &
  \multicolumn{1}{l|}{59.40\%} &
  90578.27 &
  1356.36 &
  3.94 &
  0.01 \\
 &
  \multicolumn{1}{l|}{\textit{Int-16}} &
  \multicolumn{1}{l|}{16.40\%} &
  1845.00 &
  27.57 &
  0.54 &
  \multicolumn{1}{l|}{0.01} &
  \multicolumn{1}{l|}{40.80\%} &
  16614.91 &
  247.64 &
  1.76 &
  \multicolumn{1}{l|}{0.01} &
  \multicolumn{1}{l|}{54.00\%} &
  62088.17 &
  929.52 &
  2.73 &
  0.01 \\
\multirow{-3}{*}{\textbf{Square-$L_\infty$}} &
  \multicolumn{1}{l|}{\textit{Int-8}} &
  \multicolumn{1}{l|}{82.30\%} &
  1125.94 &
  17.60 &
  0.36 &
  \multicolumn{1}{l|}{0.01} &
  \multicolumn{1}{l|}{86.00\%} &
  11173.88 &
  167.20 &
  1.23 &
  \multicolumn{1}{l|}{0.01} &
  \multicolumn{1}{l|}{78.70\%} &
  27792.02 &
  435.28 &
  1.35 &
  0.01 \\ \hline
\rowcolor[HTML]{C0C0C0} 
\multicolumn{17}{c}{\cellcolor[HTML]{C0C0C0}\textit{\textbf{BLACK-BOX}}} \\ \hline
 &
  \multicolumn{1}{l|}{\textit{Float-32}} &
  \multicolumn{1}{l|}{7.70\%} &
  3070.36 &
  8.15 &
  0.19 &
  \multicolumn{1}{l|}{0.01} &
  \multicolumn{1}{l|}{\cellcolor[HTML]{c0d4ef}91.30\%} & 
  \cellcolor[HTML]{c0d4ef}2764.59 &
  \cellcolor[HTML]{c0d4ef}3.59 &
  \cellcolor[HTML]{c0d4ef}0.02 &
  \multicolumn{1}{l|}{\cellcolor[HTML]{c0d4ef}0.00} &
  \multicolumn{1}{l|}{2.20\%} &
  150489.77 &
  2209.77 &
  7.15 &
  0.09 \\
 &
  \multicolumn{1}{l|}{\textit{Int-16}} &
  \multicolumn{1}{l|}{8.00\%} &
  2989.68 &
  9.10 &
  0.21 &
  \multicolumn{1}{l|}{0.02} &
  \multicolumn{1}{l|}{\cellcolor[HTML]{c0d4ef}91.00\%} &
  \cellcolor[HTML]{c0d4ef}2428.34 &
  \cellcolor[HTML]{c0d4ef}3.55 &
  \cellcolor[HTML]{c0d4ef}0.03 &
  \multicolumn{1}{l|}{\cellcolor[HTML]{c0d4ef}0.00} &
  \multicolumn{1}{l|}{0.00\%} &
  150347.67 &
  2684.71 &
  8.69 &
  0.11 \\
\multirow{-3}{*}{\textbf{Boundary}} &
  \multicolumn{1}{l|}{\textit{Int-8}} &
  \multicolumn{1}{l|}{6.50\%} &
  2450.47 &
  61.34 &
  1.38 &
  \multicolumn{1}{l|}{0.08} &
  \multicolumn{1}{l|}{\cellcolor[HTML]{c0d4ef}91.00\%} &
  \cellcolor[HTML]{c0d4ef}1928.49 &
  \cellcolor[HTML]{c0d4ef}20.30 &
  \cellcolor[HTML]{c0d4ef}0.16 &
  \multicolumn{1}{l|}{\cellcolor[HTML]{c0d4ef}0.00} &
  \multicolumn{1}{l|}{1.00\%} &
  144495.22 &
  7928.35 &
  25.56 &
  0.29 \\
\rowcolor[HTML]{EFEFEF} 
\cellcolor[HTML]{EFEFEF} &
  \multicolumn{1}{l|}{\cellcolor[HTML]{EFEFEF}\textit{Float-32}} &
  \multicolumn{1}{l|}{\cellcolor[HTML]{EFEFEF}\cellcolor[HTML]{EFEFEF}11.00\%} &
  \cellcolor[HTML]{EFEFEF}2413.11 &
  \cellcolor[HTML]{EFEFEF}36.95 &
  \cellcolor[HTML]{EFEFEF}0.77 &
  \multicolumn{1}{l|}{\cellcolor[HTML]{EFEFEF}0.03} &
  \multicolumn{1}{l|}{\cellcolor[HTML]{fffc9e}91.30\%} &
  \cellcolor[HTML]{fffc9e}828.96 &
  \cellcolor[HTML]{fffc9e}414.74 &
  \cellcolor[HTML]{fffc9e}2.80 &
  \multicolumn{1}{l|}{\cellcolor[HTML]{fffc9e}0.03} &
  \multicolumn{1}{l|}{\cellcolor[HTML]{EFEFEF}0.10\%} &
  145536.22 &
  1561.76 &
  5.09 &
  0.06 \\
\rowcolor[HTML]{EFEFEF} 
\cellcolor[HTML]{EFEFEF} &
  \multicolumn{1}{l|}{\cellcolor[HTML]{EFEFEF}\textit{Int-16}} &
  \multicolumn{1}{l|}{\cellcolor[HTML]{EFEFEF}22.00\%} &
  \cellcolor[HTML]{EFEFEF}2284.17 &
  \cellcolor[HTML]{EFEFEF}83.51 &
  \cellcolor[HTML]{EFEFEF}1.72 &
  \multicolumn{1}{l|}{\cellcolor[HTML]{EFEFEF}0.06} &
  \multicolumn{1}{l|}{\cellcolor[HTML]{fffc9e}91.00\%} &
  \cellcolor[HTML]{fffc9e}1368.56 &
  \cellcolor[HTML]{fffc9e}687.31 &
  \cellcolor[HTML]{fffc9e}4.77 &
  \multicolumn{1}{l|}{\cellcolor[HTML]{fffc9e}0.05} &
  \multicolumn{1}{l|}{1.00\%} &
  142126.62 &
  2374.74 &
  7.67 &
  0.07 \\
\rowcolor[HTML]{EFEFEF} 
\multirow{-3}{*}{\cellcolor[HTML]{EFEFEF}\textbf{GeoDA}} &
  \multicolumn{1}{l|}{\cellcolor[HTML]{EFEFEF}\textit{Int-8}} &
  \multicolumn{1}{l|}{\cellcolor[HTML]{EFEFEF}13.00\%} &
  \cellcolor[HTML]{EFEFEF}2193.12 &
  \cellcolor[HTML]{EFEFEF}98.67 &
  \cellcolor[HTML]{EFEFEF}2.08 &
  \multicolumn{1}{l|}{\cellcolor[HTML]{EFEFEF}0.08} &
  \multicolumn{1}{l|}{\cellcolor[HTML]{fffc9e}91.00\%} &
  \cellcolor[HTML]{fffc9e}826.04 &
  \cellcolor[HTML]{fffc9e}412.77 &
  \cellcolor[HTML]{fffc9e}2.79 &
  \multicolumn{1}{l|}{\cellcolor[HTML]{fffc9e}0.03} &
  \multicolumn{1}{l|}{\cellcolor[HTML]{FFCCC9}4.50\%} &
  \cellcolor[HTML]{FFCCC9}132656.10 &
  \cellcolor[HTML]{FFCCC9}10057.61 &
  \cellcolor[HTML]{FFCCC9}31.51 &
  \cellcolor[HTML]{FFCCC9}0.26 \\ \hline
\end{tabular}%
}
\end{table*}

\subsection{Direct Attacks} \label{sec:res_direct_attacks}
We compared the robustness of QNNs against full-precision ANNs using the gray-box and black-box attacks described in Table \ref{tab:state_of_art_attacks}. We do not include white-box attacks in this evaluation as these attacks can not be directly applied to QNNs. Results are detailed in Table \ref{tab:direct_attacks}.

\mypara{A1. Point distance to the decision boundary.} The results presented in Table \ref{tab:direct_attacks} indicate that int-8 models exhibit greater robustness than the float-32 and int-16 versions, with int-16 being more robust than float-32 models. When considering black-box attacks, we observe similar adversarial accuracies between every model bit-width; however, the same is not verified for the distortion metrics. Although the $L_0$ norm shows the number of perturbed pixels tends to decrease as precision lowers, the other metrics ($L_1$, $L_2$, and $L_\infty$) show the amount of distortion the perturbed pixels get is significantly higher. For reference, the $L_2$ distortion reported in Table \ref{tab:direct_attacks} for the Boundary attack on the Coffee int-8 QNN ($L_2 = 25.56$) is 3.57x higher than the distortion generated by the same attack for the Coffee float-32 ANN ($L_2 = 7.15$).

The fact that the distortion significantly increases for similar or higher adversarial accuracy as the precision lowers suggests that the average point distance to the decision boundary increases as precision lowers. This finding is reinforced in the analysis \textit{A5} of Section \ref{sec:res_transferability} and was previously observed in \cite{gorsline2021}. \textcite{gorsline2021} defined a geometric model proving that the average distance to the decision boundary increases as the weight precision and the input dimensionality decrease.

\begin{mdframed}[style=remarkstyle]
\mypara{Takeaway \thetakeawyacount \stepcounter{takeawyacount}.} The average point distance to the decision boundary increases as the precision of a neural network decreases, making int-8 models generally more robust to adversarial examples than int-16 and float-32 versions.
\end{mdframed}

\mypara{A2. Explosion/Vanishing of estimated gradients (gradient masking).} Quantization makes it more difficult to optimize over the loss surface. On the one hand, when the perturbation added by the attack is small, it may be insufficient to shift the quantization bucket. Consequently, the model will return the same output, making the gradient estimated by the attack null. On the other hand, when the perturbation is large enough, the output variation could be so significant that it may explode the gradient estimation. These scenarios, where the gradient becomes difficult to compute, are referred to as gradient masking.

Although gray- and black-box attacks reduce the chance of gradient masking, we can still observe some scenarios of gradient masking when we compare the results of direct attacks (Table \ref{tab:direct_attacks}) against adversarial transferability (Table \ref{tab:transferability}) for int-8 models. This behavior is noticeable in the Square and ZOO attacks on all datasets. For this set of attacks, attacking the float-32 ANN and transferring the adversarial examples to the int-8 QNNs have more impact on the adversarial accuracy than using the attacks directly on the int-8 QNNs. For instance, when directly attacking the CIFAR-10 int-8 QNN using the Square-$L_\infty$ attack, we register, in Table \ref{tab:direct_attacks}, an adversarial accuracy of 82,30\%. However, if we use adversarial example transferability to attack this same QNN, its adversarial accuracy drops to 29,90\% (Table \ref{tab:transferability}).

Supported by the collected empirical evidence, we argue that quantization makes the loss surface of models harder to optimize over as quantization increases the possibility of gradient explosion or vanishing. To prove our argument, we calculate the average zero density for the gradients estimated for the float-32, int-16, and int-8 models trained on CIFAR-10. As the quantization policy we are using does not enable the exact calculus of the gradient, we use the finite difference method to estimate it. We observed an average zero density of 70.20\%, 72.68\%, and 88.18\% for the gradients estimated in float-32, int-16, and int-8 models. As the zero density increases as the precision lowers, we can conclude that quantization increases the chance of gradient masking.

\textcite{bernhard2019} arrived at a similar conclusion. They observed that the quantization of activations could make the gradient explode or vanish. However, they did not observe the same behavior for weight-only quantized models, i.e., models with weights quantized to integer precision but whose activations remain in floating-point.

\begin{mdframed}[style=remarkstyle]
\mypara{Takeaway \thetakeawyacount \stepcounter{takeawyacount}.} Quantization increases the chance of gradient masking. Small perturbations may be erased by quantization and make the gradient vanish while bigger perturbations can make the gradient explode.
\end{mdframed}

\begin{figure*}[!t]
	\centering
	\includegraphics[width=\linewidth]{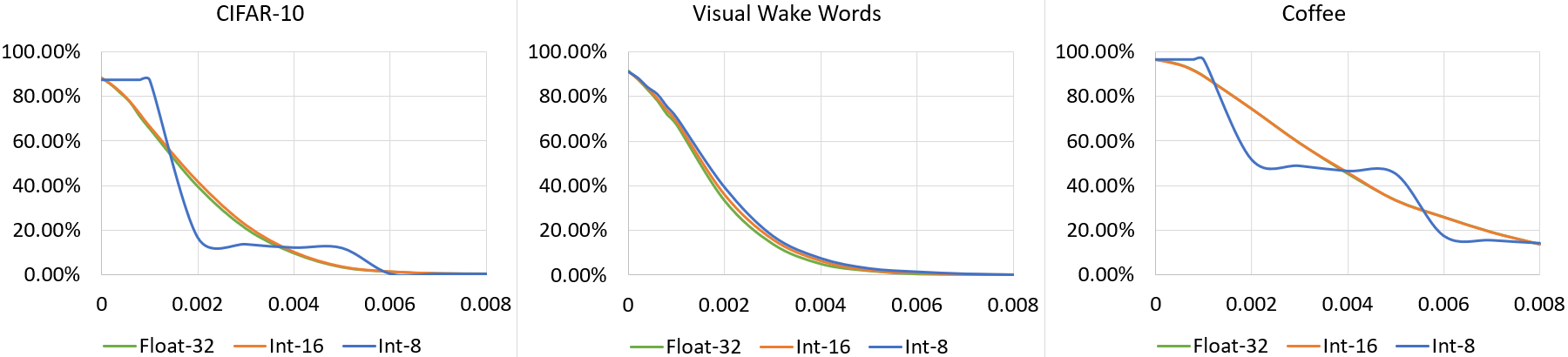}
	\caption{Accuracy \textit{vs.} $L_\infty$ distortion for PGD attack: adversarial example transferability from float-32 ANNs to int-16 and int-8 QNNs.}
	\label{fig:pgd_transferability}
\end{figure*}
\begin{table*}[t]
\caption{Quantization impact on transferability}
\label{tab:transferability}
\resizebox{\textwidth}{!}{%
\begin{tabular}{lllllllllllllllll}
\hline
\textbf{Attack} &
  \multicolumn{1}{l|}{} &
  \multicolumn{5}{c}{\textbf{CIFAR-10}} &
  \multicolumn{5}{c}{\textbf{Visual Wake Words}} &
  \multicolumn{5}{c}{\textbf{Coffee Dataset}} \\ \cline{3-17} 
 &
  \multicolumn{1}{l|}{} &
  \multicolumn{1}{c|}{\textit{Acc}} &
  \multicolumn{1}{c}{\textit{$L_0$}} &
  \multicolumn{1}{c}{\textit{$L_1$}} &
  \multicolumn{1}{c}{\textit{$L_2$}} &
  \multicolumn{1}{c|}{\textit{$L_\infty$}} &
  \multicolumn{1}{c|}{\textit{Acc}} &
  \multicolumn{1}{c}{\textit{$L_0$}} &
  \multicolumn{1}{c}{\textit{$L_1$}} &
  \multicolumn{1}{c}{\textit{$L_2$}} &
  \multicolumn{1}{c|}{\textit{$L_\infty$}} &
  \multicolumn{1}{c|}{\textit{Acc}} &
  \multicolumn{1}{c}{\textit{$L_0$}} &
  \multicolumn{1}{c}{\textit{$L_1$}} &
  \multicolumn{1}{c}{\textit{$L_2$}} &
  \multicolumn{1}{c}{\textit{$L_\infty$}} \\ \hline
\rowcolor[HTML]{C0C0C0} 
\multicolumn{17}{c}{\cellcolor[HTML]{C0C0C0}\textit{\textbf{WHITE-BOX}}} \\ \hline
 &
  \multicolumn{1}{l|}{\textit{Float-32}} &
  \multicolumn{1}{l|}{6.60\%} &
   &
   &
   &
  \multicolumn{1}{l|}{} &
  \multicolumn{1}{l|}{7.50\%} &
   &
   &
   &
  \multicolumn{1}{l|}{} &
  \multicolumn{1}{l|}{2.30\%} &
   &
   &
   &
   \\
 &
  \multicolumn{1}{l|}{\textit{Int-16}} &
  \multicolumn{1}{l|}{46.10\%} &
   &
   &
   &
  \multicolumn{1}{l|}{} &
  \multicolumn{1}{l|}{49.90\%} &
   &
   &
   &
  \multicolumn{1}{l|}{} &
  \multicolumn{1}{l|}{8.00\%} &
   &
   &
   &
   \\
\hline
\end{tabular}%
}
\end{table*}
\subsection{Transferability} \label{sec:res_transferability}
We evaluate the transferability of adversarial examples from float-32 ANNs to QNNs at different bit-widths (int-16 and int-8) using white-, gray-, and black-box attacks. Figure \ref{fig:pgd_transferability} depicts the transferability results of the PGD attack, while Table \ref{tab:transferability} details the results of the remaining attacks, configured as detailed in Table \ref{tab:attack_config} (Appendix \ref{ap:attack_defense_config}).

\mypara{A3. Transferability is sensitive to quantization-shift.} As observed in Figure \ref{fig:pgd_transferability}, for the CIFAR-10 and Coffee datasets, the lines representing the adversarial accuracy of float-32 and int-16 models overlap for the entire range of distortion. However, the same does not occur with the int-8 model. While for float-32 and int-16, the adversarial accuracy decreases smoothly as distortion increases, for the int-8 model, the adversarial accuracy decreases in buckets - it only decreases after a certain distortion threshold. We argue that this is a consequence of the quantization shift, the most widely-known effect of quantization \cite{bernhard2019, gorsline2021, zhao2018, lin2019, duncan2020, song2021}. As quantization maps more than one value on the same quantization bucket, quantization can amplify or attenuate the effect of distortion. Quantization can boost adversarial accuracy by eliminating small perturbations; however, it can amplify bigger perturbations.

Nevertheless, the effect of quantization shift depends on the target model and attack. We observe in Figure \ref{fig:pgd_transferability} that PGD achieves much better transferability results on VWW than on the other two models. Compared to the other two models, VWW is the only binary classification task. At first sight, it appears that binary models are less sensitive to the quantization-shift phenomenon. However, the results detailed in Table \ref{tab:transferability} for the VWW model clearly show this is a particular behavior of the PGD-VWW combo. In contrast to PGD, in DeepFool, JSMA, C\&W-$L_2$, EAD, and Square-$L_\infty$, the adversarial accuracy of the VWW models at different bit-widths is considerably different for the same distortion. Combining this observation with the results registered for the CIFAR-10 and Coffee models, we state that the success of transferability (transfer adversarial examples crafted on float-32 ANNs to int-16 or int-8 QNNs) is highly dependent on the attack-model combo.

\begin{mdframed}[style=remarkstyle]
\mypara{Takeaway \thetakeawyacount \stepcounter{takeawyacount}.} Adversarial example transferability is influenced by quantization-shift. Quantization can mitigate the impact of small perturbations while amplifying the effects of larger perturbations.
\end{mdframed}

\mypara{A4. Gradient misalignment.} In addition to the quantization-shift phenomenon, the poor transferability results generally observed from float-32 to int-16 and int-8 models can also be a consequence of gradient misalignment, as previously disclosed by \textcite{bernhard2019}. Quantization can map multiple float-32 values into the same int-16 or int-8 value, moving distinct ANN weights and activations into the same quantization bucket. Therefore, the gradient estimated for float-32 ANNs may differ for the equivalent int-16 or int-8 models and even have different directions, with some attacks being more affected than others by this phenomenon. 

To support our argument, we measured the average cosine similarity between the gradients of float-32, int-16, and int-8 models trained on the three target datasets. Cosine similarity calculates the cosine of the angle between two vectors, which reflects their directional similarity. As an exact calculus of the gradient is not possible for QNNs under the TFLite Micro quantization policy, we first dequantized the QNNs and converted them to float-32 ANNs. The results are detailed in Table \ref{tab:grad_cos_sml}. As observed, only the Coffee models have gradient vectors pointing in almost the same direction (cosine similarity close to 1). The cosine similarity between float-32 and int-16/int-8 bit-widths for CIFAR-10 and VWW models is 0, which means that the gradient vectors are orthogonal to each other. These results suggest that quantization causes gradient misalignment, decreasing the success of adversarial transferability between models at different bit-width.

\begin{table}[h]
\centering
\caption{Gradient cosine similarity: ANN vs. QNNs.}
\label{tab:grad_cos_sml}
\begin{tabular}{@{}l|ccc@{}}
\toprule
                  & \textbf{CIFAR-10} & \textbf{VWW} & \textbf{Coffee} \\
                  & \multicolumn{3}{c}{\textit{Float-32}}              \\ \midrule
\textit{Float-32} & 1.00              & 1.00         & 1.00            \\
\textit{Int-16}   & 0.00              & 0.00         & 0.99            \\
\textit{Int-8}    & 0.00              & 0.00         & 0.99            \\ \bottomrule
\end{tabular}
\end{table}

\begin{mdframed}[style=remarkstyle]
\mypara{Takeaway \thetakeawyacount \stepcounter{takeawyacount}.} By mapping different ANN weights and activations into the same integer value, quantization can cause gradient misalignment: the gradient estimated for float-32 ANNs may not be the same for the equivalent int-16 or int-8 models and even have different directions.
\end{mdframed}

\begin{table*}[t]
\caption{Accuracy of ANNs/QNNs enhanced with train-based defenses}
\label{tab:train_based_def_acc}
\resizebox{\textwidth}{!}{%
\begin{tabular}{@{}lcccccccccccc@{}}
\toprule
 &
  \multicolumn{4}{c}{\textbf{CIFAR-10}} &
  \multicolumn{4}{c}{\textbf{Visual Wake Words}} &
  \multicolumn{4}{c}{\textbf{Coffee Dataset}} \\ \midrule
\textbf{} &
  \multicolumn{1}{l}{\textit{Distil.}} &
  \multicolumn{1}{l}{\textit{PGD AdvT}} &
  \multicolumn{1}{l}{\textit{Ens. AdvT}} &
  \multicolumn{1}{l|}{\textit{Sink. AdvT}} &
  \multicolumn{1}{l}{\textit{Distil.}} &
  \multicolumn{1}{l}{\textit{PGD AdvT}} &
  \multicolumn{1}{l}{\textit{Ens. AdvT}} &
  \multicolumn{1}{l|}{\textit{Sink. AdT}} &
  \multicolumn{1}{l}{\textit{Distil.}} &
  \multicolumn{1}{l}{\textit{PGD AdvT}} &
  \multicolumn{1}{l}{\textit{Ens. AdvT}} &
  \multicolumn{1}{l}{\textit{Sink. AdvT}} \\ \midrule
\textbf{Float-32} &
  82.90\% &
  82.17\% &
  85.77\% &
  \multicolumn{1}{c|}{81.24\%} &
  82.91\% &
  78.72\% &
  74.85\% &
  \multicolumn{1}{c|}{77.86\%} &
  95.29\% &
  95.87\% &
  97.37\% &
  95.38\% \\
\textbf{Int-16} &
  82.77\% &
  82.21\% &
  85.85\% &
  \multicolumn{1}{c|}{81.29\%} &
  82.90\% &
  78.48\% &
  74.93\% &
  \multicolumn{1}{c|}{77.99\%} &
  95.32\% &
  95.84\% &
  97.40\% &
  95.32\% \\
\textbf{Int-8} &
  82.58\% &
  81.76\% &
  85.62\% &
  \multicolumn{1}{c|}{81.27\%} &
  82.94\% &
  78.47\% &
  74.76\% &
  \multicolumn{1}{c|}{77.86\%} &
  95.00\% &
  96.01\% &
  97.43\% &
  95.06\% \\ \bottomrule
\end{tabular}%
}
\end{table*}

\mypara{A5. Point distance to the decision boundary (A1 recall).} DeepFool determines the minimum $L_2$ distortion required to move a given sample into a different class. As a result, DeepFool is commonly used to assess the minimum $L_2$ distance to the nearest decision boundary. We observe in Table \ref{tab:transferability}, in every dataset, that for the same $L_2$ distance measured by DeepFool, the adversarial accuracy increases as the precision decreases. This behavior suggests that quantization alters the decision boundaries of neural networks, making the average point distance to the decision boundary increase as the precision of the neural network decreases. This observation supports the argument \textit{A1}.

\section{Defenses Against Adversarial Examples} \label{sec:defenses_eval}
In this section, we empirically evaluate how defenses aiming to enhance the robustness of ANNs perform when applied to int-16 and int-8 QNNs. As detailed in Table \ref{tab:state_of_art_defenses}, our empirical evaluation includes two types of defenses: (i) train-based and (ii) input pre-processing. The former implies the retraining of the model to be protected; the latter cleans the sample to remove adversarial noise before it is used as input for the neural network.

For both defense types, the evaluation is performed under white-box, gray-box, and black-box scenarios. White-box attacks are used to evaluate adversarial transferability from ANNs to QNNs, while gray-box and black-box attacks are used to attack the QNNs directly. We reduce the scope of attacks included in this part of the evaluation to the attacks that impact the most
the adversarial accuracy of QNNs before any defense:

\begin{itemize}
    \item \textbf{White-box:} We reduce the scope to DeepFool, C\&W-$L_\infty$, and AutoAttack. DeepFool is useful to determine the point distance to the decision boundary, as it returns the minimum $L_2$ distance to the closest boundary. C\&W-$L_\infty$ and AutoAttack were the only white-box attacks registering good adversarial transferability across different bit-widths.
    \item \textbf{Gray-box:} We only include the Square Attack as the results detailed in Table \ref{tab:direct_attacks} suggest that ZOO does not perform well against the neural networks we are using.
    \item \textbf{Black-box:} We include both the Boundary and GeoDA attacks; however, we do not use them against VWW models as they have failed to produce adversarial examples (Table \ref{tab:direct_attacks}). 
\end{itemize}

We consider that a defense succeeds in defending a model against an attack when the accuracy drop for the baseline test accuracy (detailed in Table \ref{tab:models_test_acc}) is below 20\%. We chose a 20\% threshold, as we believe that a model exhibiting only a 20\% decrease in accuracy under attack is notably robust. By subtracting 20\% from the baseline test accuracy of the three models under consideration (Table \ref{tab:models_test_acc}), we arrive at accuracies of 67.50\%, 71\%, and 77\% for CIFAR-10, VWW, and Coffee models, respectively. These values remain significantly higher than the accuracy of random choice for each of these models.

\subsection{Train-Based Defenses}
To compare how train-based defenses perform on QNNs, we first retrain the float-32 ANN according to the defense constraints and then quantize the resulting ANN to int-16 and int-8 to obtain the retrained QNNs. Directly retraining the QNNs is not possible as it is not possible to perform an exact calculus of the gradient. The accuracy of the retrained ANNs/QNNs is detailed in Table \ref{tab:train_based_def_acc}.

The retrained ANNs and QNNs relative to a given defense are tested similarly to vanilla ANNs and QNNs: (i) gray- and black-box attacks are applied directly to each ANN or QNN and (ii) white-box attacks are used to evaluate the transferability from float-32 ANNs to QNNs. Results are detailed in Table \ref{tab:results_train_based_defenses_compact}. For synthesis purposes, Table \ref{tab:results_train_based_defenses_compact} only reports the $L_2$ distortion. For a more comprehensive distortion report, please refer to Appendix \ref{ap:defenses_full_results}.

\begin{table}[]
\caption{Evaluation of train-based defenses. Red highlights samples that affect the accuracy and whose noise is detectable to the human eye. Green highlights samples where accuracy dropped by less than 20\%.}
\label{tab:results_train_based_defenses_compact}
\resizebox{\columnwidth}{!}{%
\begin{tabular}{llllllll}
\hline
\textbf{Attack} &
  \multicolumn{1}{l|}{} &
  \multicolumn{2}{c|}{\textbf{CIFAR-10}} &
  \multicolumn{2}{c|}{\textbf{Visual Wake Words}} &
  \multicolumn{2}{c}{\textbf{Coffee Dataset}} \\ \cline{3-8} 
 &
  \multicolumn{1}{l|}{} &
  \multicolumn{1}{c}{\textit{Acc}} &
  \multicolumn{1}{c|}{\textit{$L_2$}} &
  \multicolumn{1}{c}{\textit{Acc}} &
  \multicolumn{1}{c|}{\textit{$L_2$}} &
  \multicolumn{1}{c}{\textit{Acc}} &
  \multicolumn{1}{c}{\textit{$L_2$}} \\ \hline
\rowcolor[HTML]{C0C0C0} 
\multicolumn{8}{c}{\cellcolor[HTML]{C0C0C0}\textit{\textbf{DEFENSIVE DISTILATION}}} \\ \hline
\rowcolor[HTML]{D5D5D5} 
\multicolumn{8}{c}{\cellcolor[HTML]{D5D5D5}\textit{TRANSFERABILITY}} \\ \hline
 &
  \multicolumn{1}{l|}{\textit{Float-32}} &
  8.60\% &
  \multicolumn{1}{l|}{} &
  14.30\% &
  \multicolumn{1}{l|}{} &
  3.50\% &
   \\
 &
  \multicolumn{1}{l|}{\textit{Int-16}} &
  47.40\% &
  \multicolumn{1}{l|}{} &
  70.20\% &
  \multicolumn{1}{l|}{} &
  12.10\% &
   \\
\multirow{-3}{*}{\textbf{DeepFool}} &
  \multicolumn{1}{l|}{\textit{Int-8}} &
  \cellcolor[HTML]{CAE6CA}75.80\% &
  \multicolumn{1}{l|}{\multirow{-3}{*}{0.09}} &
  \cellcolor[HTML]{CAE6CA}78.10\% &
  \multicolumn{1}{l|}{\multirow{-3}{*}{0.19}} &
  49.40\% &
  \multirow{-3}{*}{0.61} \\
\rowcolor[HTML]{EFEFEF} 
\cellcolor[HTML]{EFEFEF} &
  \multicolumn{1}{l|}{\cellcolor[HTML]{EFEFEF}\textit{Float-32}} &
  0.00\% &
  \multicolumn{1}{l|}{\cellcolor[HTML]{EFEFEF}} &
  0.00\% &
  \multicolumn{1}{l|}{\cellcolor[HTML]{EFEFEF}} &
  0.00\% &
  \cellcolor[HTML]{EFEFEF} \\
\hline
\end{tabular}%
}
\end{table}

\mypara{D1. Point distance to the decision boundary.} To analyze the effect of the evaluated defenses on the average point distance to the decision boundary, we rely on DeepFool, as it determines the $L_2$ distance to the closest decision boundary. When we compare the results of DeepFool for models without any defense (Table \ref{tab:transferability}) against defense-enhanced models (Table \ref{tab:results_train_based_defenses_compact}), we verify that the $L_2$ norm only increased consistently across the three models for the PGD Adversarial Training and Sinkhorn Adversarial Training, with a little advantage for the later defense. For reference, the most significant increase was registered for the VWW models, where the average point distance to the decision boundary increased from $L_2=0.22$ (Table \ref{tab:transferability}) to $L_2=2.88$ (Table \ref{tab:results_train_based_defenses_compact}). In turn, Defensive Distillation only increased the average point distance to the decision boundary of CIFAR-10 models and by a slimmer margin ($L_2 = 0.08$ vs. $L_2 = 0.09$), while Ensemble Adversarial Training only for Coffee ($L_2 = 0.80$ vs. $L_2 = 0.85$).

Interestingly, int-8 models enhanced with PGD and Sinkhorn Adversarial Training present a lower adversarial accuracy than those enhanced with the other defenses. This might suggest that despite PGD and Sinkhorn Adversarial Training increasing the average point distance to the decision boundary, this feature does not hold after quantization. However, the reduced adversarial accuracy of int-8 models enhanced with PGD and Sinkhorn Adversarial Training occurs because of the quantization-shift phenomenon. As these defenses significantly increase the average point distance to the decision boundary, DeepFool returns a higher perturbation that can move the quantized adversarial examples to the next quantization bucket and have a higher impact on the adversarial accuracy of int-8 and int-16 models.

\begin{mdframed}[style=remarkstyle]
\mypara{Takeaway \thetakeawyacount \stepcounter{takeawyacount}.} PGD and Sinkhorn Adversarial Training consistently increase the average point distance to the decision boundary, and this feature holds after quantization. Defensive Distillation and Ensemble Adversarial Training are not consistent in this topic.
\end{mdframed}

\mypara{D2. Adversarial example transferability.} The train-based defenses considered in this study do not significantly affect the transferability of adversarial examples from float-32 ANNs to int-16 or int-8 QNNs. Without any defense in place, we observe in Table \ref{tab:transferability} that AutoAttack and C\&W-$L_\infty$ attacks hold good transferability results across all bit-widths, registering a maximum accuracy loss of 5.60\% registered for C\&W-$L_\infty$ attack on CIFAR-10. With defense in place, the maximum adversarial accuracy loss when transferring adversarial examples crafted on these attacks rises to 17.20\%. This value is registered for the VWW models enhanced with PGD Adversarial Training when attacked by C\&W-$L_\infty$ (Table \ref{tab:results_train_based_defenses_compact}).

This behavior is verified even for Ensemble Adversarial Training, specifically developed to counteract adversarial example transferability. However, after empirical validation, we can attest that this defense is not ready to handle adversarial example transferability between models at different bit widths. The defense must address the quantization-shift and gradient misalignment phenomenon to counteract adversarial example transferability adequately. To the best of our knowledge, no existing defense has accounted for these factors.

\begin{mdframed}[style=remarkstyle]
\mypara{Takeaway \thetakeawyacount \stepcounter{takeawyacount}.} None of the training-based defenses mitigate adversarial example transferability. We envision this is a consequence of none of them smoothing the quantization-shift and gradient misalignment, which are required to tackle the transferability.
\end{mdframed}

\mypara{D3. Defense transferability.} We observe that the robustness guarantees hold after quantization: whenever the defense is robust to protect a float-32 ANN, it is also robust in the int-16 and int-8 QNNs. However, the opposite is not valid. Let us consider the effect of Defensive Distillation against the Square-$L_\infty$ attack. We observe that the defense delivers solid results across the three float-32 ANNs used for evaluation, and this behavior extends to the lower QNN's bit widths. However, let us consider the results of the same defense in the face of the Square-$L_2$ attack. We observe that Defensive Distillation greatly enhances the robustness of int-8 QNNs, but the behavior does not apply to the higher bit widths. This results from the combined robustness of quantization and the defense itself.

\begin{mdframed}[style=remarkstyle]
\mypara{Takeaway \thetakeawyacount \stepcounter{takeawyacount}.} Every time a defense is effective in a full-precision ANN, it is likewise effective in the corresponding QNNs. A defense may, however, be adequate to defend a low-precision QNN but insufficient to defend a full-precision ANN.
\end{mdframed}

\subsection{Pre-Processing Defenses}
\begin{table}[]
\centering
\caption{Accuracy of ANNs/QNNs enhanced with pre-processing defenses}
\label{tab:pre_proc_def_acc}
\begin{tabular}{@{}lcc|c|c@{}}
\toprule
                  & \multicolumn{2}{c|}{\textbf{CIFAR-10}} & \textbf{VWW} & \textbf{Coffee} \\ \midrule
\textbf{} &
  \multicolumn{1}{l}{\textit{Feat. Sq.}} &
  \multicolumn{1}{l|}{\textit{Pixel Defend}} &
  \multicolumn{1}{l|}{\textit{Feat. Sq.}} &
  \multicolumn{1}{l}{\textit{Feat. Sq.}} \\ \midrule
\textbf{Float-32} & 83.70\%            & 84.80\%           & 97.00\%      & 96.80\%         \\
\textbf{Int-16}   & 83.60\%            & 85.10\%           & 96.90\%      & 97.00\%         \\
\textbf{Int-8}    & 83.00\%            & 84.90\%           & 96.60\%      & 96.90\%         \\ \bottomrule
\end{tabular}
\end{table}

\begin{table*}[]
\centering
\caption{Evaluation of Feature Squeezing. Green highlights samples where accuracy dropped by less than 20\%. For CIFAR-10 and VWW the defense succeeded for every attack and bit-width tested.}
\label{tab:results_feature_squeezing}
\resizebox{\textwidth}{!}{%
\begin{tabular}{lllllllllllllllll}
\hline
\textbf{Attack} &
  \multicolumn{1}{l|}{} &
  \multicolumn{5}{c}{\textbf{\cellcolor[HTML]{CAE6CA}CIFAR-10}} &
  \multicolumn{5}{c}{\textbf{\cellcolor[HTML]{CAE6CA}Visual Wake Words}} &
  \multicolumn{5}{c}{\textbf{Coffee Dataset}} \\ \cline{3-17} 
 &
  \multicolumn{1}{l|}{} &
  \multicolumn{1}{c|}{\textit{Acc}} &
  \multicolumn{1}{c}{\textit{$L_0$}} &
  \multicolumn{1}{c}{\textit{$L_1$}} &
  \multicolumn{1}{c}{\textit{$L_2$}} &
  \multicolumn{1}{c|}{\textit{$L_\infty$}} &
  \multicolumn{1}{c|}{\textit{Acc}} &
  \multicolumn{1}{c}{\textit{$L_0$}} &
  \multicolumn{1}{c}{\textit{$L_1$}} &
  \multicolumn{1}{c}{\textit{$L_2$}} &
  \multicolumn{1}{c|}{\textit{$L_\infty$}} &
  \multicolumn{1}{c|}{\textit{Acc}} &
  \multicolumn{1}{c}{\textit{$L_0$}} &
  \multicolumn{1}{c}{\textit{$L_1$}} &
  \multicolumn{1}{c}{\textit{$L_2$}} &
  \multicolumn{1}{c}{\textit{$L_\infty$}} \\ \hline
\rowcolor[HTML]{C0C0C0} 
\multicolumn{17}{c}{\cellcolor[HTML]{C0C0C0}\textit{\textbf{FEATURE SQUEEZING}}} \\ \hline
\rowcolor[HTML]{D5D5D5} 
\multicolumn{17}{c}{\cellcolor[HTML]{D5D5D5}\textit{TRANSFERABILITY}} \\ \hline
 &
  \multicolumn{1}{l|}{\textit{Float-32}} &
  \multicolumn{1}{l|}{81.40\%} &
   &
   &
   &
  \multicolumn{1}{l|}{} &
  \multicolumn{1}{l|}{97.00\%} &
   &
   &
   &
  \multicolumn{1}{l|}{} &
  \multicolumn{1}{l|}{53.90\%} &
   &
   &
   &
   \\
 &
  \multicolumn{1}{l|}{\textit{Int-16}} &
  \multicolumn{1}{l|}{81.30\%} &
   &
   &
   &
  \multicolumn{1}{l|}{} &
  \multicolumn{1}{l|}{97.00\%} &
   &
   &
   &
  \multicolumn{1}{l|}{} &
  \multicolumn{1}{l|}{54.00\%} &
   &
   &
   &
   \\
\multirow{-3}{*}{\textbf{DeepFool}} &
  \multicolumn{1}{l|}{\textit{Int-8}} &
  \multicolumn{1}{l|}{81.10\%} &
  \multirow{-3}{*}{3056.59} &
  \multirow{-3}{*}{3.20} &
  \multirow{-3}{*}{0.08} &
  \multicolumn{1}{l|}{\multirow{-3}{*}{0.01}} &
  \multicolumn{1}{l|}{96.90\%} &
  \multirow{-3}{*}{27552.32} &
  \multirow{-3}{*}{22.88} &
  \multirow{-3}{*}{0.22} &
  \multicolumn{1}{l|}{\multirow{-3}{*}{0.02}} &
  \multicolumn{1}{l|}{63.50\%} &
  \multirow{-3}{*}{123801.09} &
  \multirow{-3}{*}{124.11} &
  \multirow{-3}{*}{0.80} &
  \multirow{-3}{*}{0.04} \\
\rowcolor[HTML]{EFEFEF} 
\cellcolor[HTML]{EFEFEF} &
  \multicolumn{1}{l|}{\cellcolor[HTML]{EFEFEF}\textit{Float-32}} &
  \multicolumn{1}{l|}{\cellcolor[HTML]{EFEFEF}70.50\%} &
  \cellcolor[HTML]{EFEFEF} &
  \cellcolor[HTML]{EFEFEF} &
  \cellcolor[HTML]{EFEFEF} &
  \multicolumn{1}{l|}{\cellcolor[HTML]{EFEFEF}} &
  \multicolumn{1}{l|}{\cellcolor[HTML]{EFEFEF}94.30\%} &
  \cellcolor[HTML]{EFEFEF} &
  \cellcolor[HTML]{EFEFEF} &
  \cellcolor[HTML]{EFEFEF} &
  \multicolumn{1}{l|}{\cellcolor[HTML]{EFEFEF}} &
  \multicolumn{1}{l|}{\cellcolor[HTML]{EFEFEF}13.60\%} &
  \cellcolor[HTML]{EFEFEF} &
  \cellcolor[HTML]{EFEFEF} &
  \cellcolor[HTML]{EFEFEF} &
  \cellcolor[HTML]{EFEFEF} \\
\rowcolor[HTML]{EFEFEF} 
\cellcolor[HTML]{EFEFEF} &
  \multicolumn{1}{l|}{\cellcolor[HTML]{EFEFEF}\textit{Int-16}} &
  \multicolumn{1}{l|}{\cellcolor[HTML]{EFEFEF}70.40\%} &
  \cellcolor[HTML]{EFEFEF} &
  \cellcolor[HTML]{EFEFEF} &
  \cellcolor[HTML]{EFEFEF} &
  \multicolumn{1}{l|}{\cellcolor[HTML]{EFEFEF}} &
  \multicolumn{1}{l|}{\cellcolor[HTML]{EFEFEF}93.30\%} &
  \cellcolor[HTML]{EFEFEF} &
  \cellcolor[HTML]{EFEFEF} &
  \cellcolor[HTML]{EFEFEF} &
  \multicolumn{1}{l|}{\cellcolor[HTML]{EFEFEF}} &
  \multicolumn{1}{l|}{\cellcolor[HTML]{EFEFEF}13.40\%} &
  \cellcolor[HTML]{EFEFEF} &
  \cellcolor[HTML]{EFEFEF} &
  \cellcolor[HTML]{EFEFEF} &
  \cellcolor[HTML]{EFEFEF} \\
\rowcolor[HTML]{EFEFEF} 
\multirow{-3}{*}{\cellcolor[HTML]{EFEFEF}\textbf{C\&W-$L_\infty$}} &
  \multicolumn{1}{l|}{\cellcolor[HTML]{EFEFEF}\textit{Int-8}} &
  \multicolumn{1}{l|}{\cellcolor[HTML]{EFEFEF}71.40\%} &
  \multirow{-3}{*}{\cellcolor[HTML]{EFEFEF}3072.00} &
  \multirow{-3}{*}{\cellcolor[HTML]{EFEFEF}13.40} &
  \multirow{-3}{*}{\cellcolor[HTML]{EFEFEF}0.26} &
  \multicolumn{1}{l|}{\multirow{-3}{*}{\cellcolor[HTML]{EFEFEF}0.01}} &
  \multicolumn{1}{l|}{\cellcolor[HTML]{EFEFEF}93.70\%} &
  \multirow{-3}{*}{\cellcolor[HTML]{EFEFEF}27647.93} &
  \multirow{-3}{*}{\cellcolor[HTML]{EFEFEF}121.68} &
  \multirow{-3}{*}{\cellcolor[HTML]{EFEFEF}0.81} &
  \multicolumn{1}{l|}{\multirow{-3}{*}{\cellcolor[HTML]{EFEFEF}0.01}} &
  \multicolumn{1}{l|}{\cellcolor[HTML]{EFEFEF}15.20\%} &
  \multirow{-3}{*}{\cellcolor[HTML]{EFEFEF}148330.23} &
  \multirow{-3}{*}{\cellcolor[HTML]{EFEFEF}765.26} &
  \multirow{-3}{*}{\cellcolor[HTML]{EFEFEF}2.42} &
  \multirow{-3}{*}{\cellcolor[HTML]{EFEFEF}0.01} \\
 &
  \multicolumn{1}{l|}{\textit{Float-32}} &
  \multicolumn{1}{l|}{70.10\%} &
   &
   &
   &
  \multicolumn{1}{l|}{} &
  \multicolumn{1}{l|}{95.60\%} &
   &
   &
   &
  \multicolumn{1}{l|}{} &
  \multicolumn{1}{l|}{9.60\%} &
   &
   &
   &
   \\
 &
  \multicolumn{1}{l|}{\textit{Int-16}} &
  \multicolumn{1}{l|}{70.70\%} &
   &
   &
   &
  \multicolumn{1}{l|}{} &
  \multicolumn{1}{l|}{94.80\%} &
   &
   &
   &
  \multicolumn{1}{l|}{} &
  \multicolumn{1}{l|}{9.60\%} &
   &
   &
   &
   \\
\multirow{-3}{*}{\textbf{AutoAttack}} &
  \multicolumn{1}{l|}{\textit{Int-8}} &
  \multicolumn{1}{l|}{70.60\%} &
  \multirow{-3}{*}{3052.49} &
  \multirow{-3}{*}{9.35} &
  \multirow{-3}{*}{0.17} &
  \multicolumn{1}{l|}{\multirow{-3}{*}{0.00}} &
  \multicolumn{1}{l|}{95.30\%} &
  \multirow{-3}{*}{24522.23} &
  \multirow{-3}{*}{91.38} &
  \multirow{-3}{*}{0.56} &
  \multicolumn{1}{l|}{\multirow{-3}{*}{0.00}} &
  \multicolumn{1}{l|}{14.40\%} &
  \multirow{-3}{*}{136814.84} &
  \multirow{-3}{*}{1065.17} &
  \multirow{-3}{*}{2.87} &
  \multirow{-3}{*}{0.01} \\ \hline
\rowcolor[HTML]{D5D5D5} 
\multicolumn{17}{c}{\cellcolor[HTML]{D5D5D5}\textit{DIRECT ATTACK}} \\ \hline
 &
  \multicolumn{1}{l|}{\textit{Float-32}} &
  \multicolumn{1}{l|}{69.40\%} &
  3044.04 &
  14.75 &
  0.87 &
  \multicolumn{1}{l|}{0.22} &
  \multicolumn{1}{l|}{93.40\%} &
  12721.11 &
  98.78 &
  1.80 &
  \multicolumn{1}{l|}{0.28} &
  \multicolumn{1}{l|}{50.50\%} &
  74699.49 &
  218.60 &
  1.80 &
  0.22 \\
 &
  \multicolumn{1}{l|}{\textit{Int-16}} &
  \multicolumn{1}{l|}{71.00\%} &
  1139.35 &
  14.01 &
  0.83 &
  \multicolumn{1}{l|}{0.21} &
  \multicolumn{1}{l|}{92.00\%} &
  12244.77 &
  216.54 &
  2.52 &
  \multicolumn{1}{l|}{0.28} &
  \multicolumn{1}{l|}{53.00\%} &
  64190.22 &
  2715.53 &
  8.57 &
  0.22 \\
\multirow{-3}{*}{\textbf{Square-$L_2$}} &
  \multicolumn{1}{l|}{\textit{Int-8}} &
  \multicolumn{1}{l|}{73.80\%} &
  443.77 &
  11.44 &
  0.75 &
  \multicolumn{1}{l|}{0.22} &
  \multicolumn{1}{l|}{94.00\%} &
  7176.22 &
  352.30 &
  3.64 &
  \multicolumn{1}{l|}{0.32} &
  \multicolumn{1}{l|}{68.80\%} &
  16830.86 &
  627.71 &
  3.11 &
  0.23 \\
\rowcolor[HTML]{EFEFEF} 
\cellcolor[HTML]{EFEFEF} &
  \multicolumn{1}{l|}{\cellcolor[HTML]{EFEFEF}\textit{Float-32}} &
  \multicolumn{1}{l|}{\cellcolor[HTML]{EFEFEF}76.30\%} &
  3045.62 &
  33.26 &
  0.65 &
  \multicolumn{1}{l|}{\cellcolor[HTML]{EFEFEF}0.01} &
  \multicolumn{1}{l|}{\cellcolor[HTML]{EFEFEF}96.00\%} &
  16671.67 &
  248.55 &
  1.78 &
  \multicolumn{1}{l|}{\cellcolor[HTML]{EFEFEF}0.01} &
  \multicolumn{1}{l|}{\cellcolor[HTML]{CAE6CA}77.20\%} &
  \cellcolor[HTML]{CAE6CA}90578.27 &
  \cellcolor[HTML]{CAE6CA}1356.36 &
  \cellcolor[HTML]{CAE6CA}3.94 &
  \cellcolor[HTML]{CAE6CA}0.01 \\
\rowcolor[HTML]{EFEFEF} 
\cellcolor[HTML]{EFEFEF} &
  \multicolumn{1}{l|}{\cellcolor[HTML]{EFEFEF}\textit{Int-16}} &
  \multicolumn{1}{l|}{\cellcolor[HTML]{EFEFEF}78.00\%} &
  1845.00 &
  27.57 &
  0.54 &
  \multicolumn{1}{l|}{\cellcolor[HTML]{EFEFEF}0.01} &
  \multicolumn{1}{l|}{\cellcolor[HTML]{EFEFEF}96.20\%} &
  16614.91 &
  247.64 &
  1.76 &
  \multicolumn{1}{l|}{\cellcolor[HTML]{EFEFEF}0.01} &
  \multicolumn{1}{l|}{\cellcolor[HTML]{EFEFEF}69.00\%} &
  62088.17 &
  929.52 &
  2.73 &
  0.01 \\
\rowcolor[HTML]{EFEFEF} 
\multirow{-3}{*}{\cellcolor[HTML]{EFEFEF}\textbf{Square-$L_\infty$}} &
  \multicolumn{1}{l|}{\cellcolor[HTML]{EFEFEF}\textit{Int-8}} &
  \multicolumn{1}{l|}{\cellcolor[HTML]{EFEFEF}80.40\%} &
  1125.94 &
  17.60 &
  0.36 &
  \multicolumn{1}{l|}{\cellcolor[HTML]{EFEFEF}0.01} &
  \multicolumn{1}{l|}{\cellcolor[HTML]{EFEFEF}95.80\%} &
  11173.88 &
  167.20 &
  1.23 &
  \multicolumn{1}{l|}{\cellcolor[HTML]{EFEFEF}0.01} &
  \multicolumn{1}{l|}{\cellcolor[HTML]{CAE6CA}82.40\%} &
  \cellcolor[HTML]{CAE6CA}27792.02 &
  \cellcolor[HTML]{CAE6CA}435.28 &
  \cellcolor[HTML]{CAE6CA}1.35 &
  \cellcolor[HTML]{CAE6CA}0.01 \\
 &
  \multicolumn{1}{l|}{\textit{Float-32}} &
  \multicolumn{1}{l|}{80.90\%} &
  3070.36 &
  8.15 &
  0.19 &
  \multicolumn{1}{l|}{0.01} &
  \multicolumn{1}{l|}{-} &
  - &
  - &
  - &
  \multicolumn{1}{l|}{-} &
  \multicolumn{1}{l|}{\cellcolor[HTML]{CAE6CA}84.30\%} &
  \cellcolor[HTML]{CAE6CA}150489.77 &
  \cellcolor[HTML]{CAE6CA}2209.77 &
  \cellcolor[HTML]{CAE6CA}7.15 &
  \cellcolor[HTML]{CAE6CA}0.09 \\
 &
  \multicolumn{1}{l|}{\textit{Int-16}} &
  \multicolumn{1}{l|}{85.00\%} &
  2989.68 &
  9.10 &
  0.21 &
  \multicolumn{1}{l|}{0.02} &
  \multicolumn{1}{l|}{-} &
  - &
  - &
  - &
  \multicolumn{1}{l|}{-} &
  \multicolumn{1}{l|}{\cellcolor[HTML]{CAE6CA}83.00\%} &
  \cellcolor[HTML]{CAE6CA}150347.67 &
  \cellcolor[HTML]{CAE6CA}2684.71 &
  \cellcolor[HTML]{CAE6CA}8.69 &
  \cellcolor[HTML]{CAE6CA}0.11 \\
\multirow{-3}{*}{\textbf{Boundary}} &
  \multicolumn{1}{l|}{\textit{Int-8}} &
  \multicolumn{1}{l|}{82.50\%} &
  2450.47 &
  61.34 &
  1.38 &
  \multicolumn{1}{l|}{0.08} &
  \multicolumn{1}{l|}{-} &
  - &
  - &
  - &
  \multicolumn{1}{l|}{-} &
  \multicolumn{1}{l|}{75.50\%} &
  144495.22 &
  7928.35 &
  25.56 &
  0.29 \\
\rowcolor[HTML]{EFEFEF} 
\cellcolor[HTML]{EFEFEF} &
  \multicolumn{1}{l|}{\cellcolor[HTML]{EFEFEF}\textit{Float-32}} &
  \multicolumn{1}{l|}{\cellcolor[HTML]{EFEFEF}81.20\%} &
  2413.11 &
  36.95 &
  0.77 &
  \multicolumn{1}{l|}{\cellcolor[HTML]{EFEFEF}0.03} &
  \multicolumn{1}{l|}{\cellcolor[HTML]{EFEFEF}-} &
  - &
  - &
  - &
  \multicolumn{1}{l|}{\cellcolor[HTML]{EFEFEF}-} &
  \multicolumn{1}{l|}{\cellcolor[HTML]{EFEFEF}50.80\%} &
  145536.22 &
  1561.76 &
  5.09 &
  0.06 \\
\rowcolor[HTML]{EFEFEF} 
\cellcolor[HTML]{EFEFEF} &
  \multicolumn{1}{l|}{\cellcolor[HTML]{EFEFEF}\textit{Int-16}} &
  \multicolumn{1}{l|}{\cellcolor[HTML]{EFEFEF}85.00\%} &
  2284.17 &
  83.51 &
  1.72 &
  \multicolumn{1}{l|}{\cellcolor[HTML]{EFEFEF}0.06} &
  \multicolumn{1}{l|}{\cellcolor[HTML]{EFEFEF}-} &
  - &
  - &
  - &
  \multicolumn{1}{l|}{\cellcolor[HTML]{EFEFEF}-} &
  \multicolumn{1}{l|}{\cellcolor[HTML]{EFEFEF}61.00\%} &
  142126.62 &
  2374.74 &
  7.67 &
  0.07 \\
\rowcolor[HTML]{EFEFEF} 
\multirow{-3}{*}{\cellcolor[HTML]{EFEFEF}\textbf{GeoDA}} &
  \multicolumn{1}{l|}{\cellcolor[HTML]{EFEFEF}\textit{Int-8}} &
  \multicolumn{1}{l|}{\cellcolor[HTML]{EFEFEF}75.00\%} &
  2193.12 &
  98.67 &
  2.08 &
  \multicolumn{1}{l|}{\cellcolor[HTML]{EFEFEF}0.08} &
  \multicolumn{1}{l|}{\cellcolor[HTML]{EFEFEF}-} &
  - &
  - &
  - &
  \multicolumn{1}{l|}{\cellcolor[HTML]{EFEFEF}-} &
  \multicolumn{1}{l|}{\cellcolor[HTML]{EFEFEF}50.00\%} &
  132656.10 &
  10057.61 &
  31.51 &
  0.26 \\ \hline
\end{tabular}%
}
\end{table*}
\begin{table}[]
\centering
\caption{Evaluation of Pixel Defend}
\label{tab:pixel_defend}
\resizebox{\columnwidth}{!}{%
\begin{tabular}{lllllll}
\hline
\textbf{Attack} &
  \multicolumn{1}{l|}{} &
  \multicolumn{5}{c}{\textbf{CIFAR-10}} \\ \cline{3-7} 
 &
  \multicolumn{1}{l|}{} &
  \multicolumn{1}{c|}{\textit{Acc}} &
  \multicolumn{1}{c}{\textit{$L_0$}} &
  \multicolumn{1}{c}{\textit{$L_1$}} &
  \multicolumn{1}{c}{\textit{$L_2$}} &
  \multicolumn{1}{c}{\textit{$L_\infty$}} \\ \hline
\rowcolor[HTML]{C0C0C0} 
\multicolumn{7}{c}{\cellcolor[HTML]{C0C0C0}\textit{\textbf{PIXEL DEFEND}}} \\ \hline
\rowcolor[HTML]{D5D5D5} 
\multicolumn{7}{c}{\cellcolor[HTML]{D5D5D5}\textit{TRANSFERABILITY}} \\ \hline
 &
  \multicolumn{1}{l|}{\textit{Float-32}} &
  \multicolumn{1}{l|}{\cellcolor[HTML]{CAE6CA}77.20\%} &
  \cellcolor[HTML]{CAE6CA} &
  \cellcolor[HTML]{CAE6CA} &
  \cellcolor[HTML]{CAE6CA} &
  \cellcolor[HTML]{CAE6CA} \\
 &
  \multicolumn{1}{l|}{\textit{Int-16}} &
  \multicolumn{1}{l|}{\cellcolor[HTML]{CAE6CA}77.40\%} &
  \cellcolor[HTML]{CAE6CA} &
  \cellcolor[HTML]{CAE6CA} &
  \cellcolor[HTML]{CAE6CA} &
  \cellcolor[HTML]{CAE6CA} \\
\multirow{-3}{*}{\textbf{DeepFool}} &
  \multicolumn{1}{l|}{\textit{Int-8}} &
  \multicolumn{1}{l|}{\cellcolor[HTML]{CAE6CA}83.20\%} &
  \multirow{-3}{*}{\cellcolor[HTML]{CAE6CA}3056.59} &
  \multirow{-3}{*}{\cellcolor[HTML]{CAE6CA}3.20} &
  \multirow{-3}{*}{\cellcolor[HTML]{CAE6CA}0.08} &
  \multirow{-3}{*}{\cellcolor[HTML]{CAE6CA}0.01} \\
\rowcolor[HTML]{EFEFEF} 
\cellcolor[HTML]{EFEFEF} &
  \multicolumn{1}{l|}{\cellcolor[HTML]{EFEFEF}\textit{Float-32}} &
  \multicolumn{1}{l|}{\cellcolor[HTML]{EFEFEF}53.90\%} &
  \cellcolor[HTML]{EFEFEF} &
  \cellcolor[HTML]{EFEFEF} &
  \cellcolor[HTML]{EFEFEF} &
  \cellcolor[HTML]{EFEFEF} \\
\rowcolor[HTML]{EFEFEF} 
\cellcolor[HTML]{EFEFEF} &
  \multicolumn{1}{l|}{\cellcolor[HTML]{EFEFEF}\textit{Int-16}} &
  \multicolumn{1}{l|}{\cellcolor[HTML]{EFEFEF}53.80\%} &
  \cellcolor[HTML]{EFEFEF} &
  \cellcolor[HTML]{EFEFEF} &
  \cellcolor[HTML]{EFEFEF} &
  \cellcolor[HTML]{EFEFEF} \\
\rowcolor[HTML]{EFEFEF} 
\multirow{-3}{*}{\cellcolor[HTML]{EFEFEF}\textbf{C\&W-$L_\infty$}} &
  \multicolumn{1}{l|}{\cellcolor[HTML]{EFEFEF}\textit{Int-8}} &
  \multicolumn{1}{l|}{\cellcolor[HTML]{EFEFEF}58.60\%} &
  \multirow{-3}{*}{\cellcolor[HTML]{EFEFEF}3072.00} &
  \multirow{-3}{*}{\cellcolor[HTML]{EFEFEF}13.40} &
  \multirow{-3}{*}{\cellcolor[HTML]{EFEFEF}0.26} &
  \multirow{-3}{*}{\cellcolor[HTML]{EFEFEF}0.01} \\
 &
  \multicolumn{1}{l|}{\textit{Float-32}} &
  \multicolumn{1}{l|}{39.40\%} &
   &
   &
   &
   \\
 &
  \multicolumn{1}{l|}{\textit{Int-16}} &
  \multicolumn{1}{l|}{39.60\%} &
   &
   &
   &
   \\
\multirow{-3}{*}{\textbf{AutoAttack}} &
  \multicolumn{1}{l|}{\textit{Int-8}} &
  \multicolumn{1}{l|}{55.20\%} &
  \multirow{-3}{*}{3052.49} &
  \multirow{-3}{*}{9.35} &
  \multirow{-3}{*}{0.17} &
  \multirow{-3}{*}{0.00} \\ \hline
\rowcolor[HTML]{D5D5D5} 
\multicolumn{7}{c}{\cellcolor[HTML]{D5D5D5}\textit{DIRECT ATTACK}} \\ \hline
 &
  \multicolumn{1}{l|}{\textit{Float-32}} &
  \multicolumn{1}{l|}{65.30\%} &
  3044.04 &
  14.75 &
  0.87 &
  0.22 \\
 &
  \multicolumn{1}{l|}{\textit{Int-16}} &
  \multicolumn{1}{l|}{66.40\%} &
  1139.35 &
  14.01 &
  0.83 &
  0.21 \\
\multirow{-3}{*}{\textbf{Square-$L_2$}} &
  \multicolumn{1}{l|}{\textit{Int-8}} &
  \multicolumn{1}{l|}{\cellcolor[HTML]{CAE6CA}71.70\%} &
  \cellcolor[HTML]{CAE6CA}443.77 &
  \cellcolor[HTML]{CAE6CA}11.44 &
  \cellcolor[HTML]{CAE6CA}0.75 &
  \cellcolor[HTML]{CAE6CA}0.22 \\
\rowcolor[HTML]{EFEFEF} 
\cellcolor[HTML]{EFEFEF} &
  \multicolumn{1}{l|}{\cellcolor[HTML]{EFEFEF}\textit{Float-32}} &
  \multicolumn{1}{l|}{\cellcolor[HTML]{EFEFEF}\cellcolor[HTML]{CAE6CA}75.80\%} &
  \cellcolor[HTML]{CAE6CA}3045.62 &
  \cellcolor[HTML]{CAE6CA}33.26 &
  \cellcolor[HTML]{CAE6CA}0.65 &
  \cellcolor[HTML]{CAE6CA}0.01 \\
\rowcolor[HTML]{EFEFEF} 
\cellcolor[HTML]{EFEFEF} &
  \multicolumn{1}{l|}{\cellcolor[HTML]{EFEFEF}\textit{Int-16}} &
  \multicolumn{1}{l|}{\cellcolor[HTML]{EFEFEF}\cellcolor[HTML]{CAE6CA}76.60\%} &
  \cellcolor[HTML]{CAE6CA}1845.00 &
  \cellcolor[HTML]{CAE6CA}27.57 &
  \cellcolor[HTML]{CAE6CA}0.54 &
  \cellcolor[HTML]{CAE6CA}0.01 \\
\rowcolor[HTML]{EFEFEF} 
\multirow{-3}{*}{\cellcolor[HTML]{EFEFEF}\textbf{Square-$L_\infty$}} &
  \multicolumn{1}{l|}{\cellcolor[HTML]{EFEFEF}\textit{Int-8}} &
  \multicolumn{1}{l|}{\cellcolor[HTML]{CAE6CA}81.20\%} &
  \cellcolor[HTML]{CAE6CA}1125.94 &
  \cellcolor[HTML]{CAE6CA}17.60 &
  \cellcolor[HTML]{CAE6CA}0.36 &
  \cellcolor[HTML]{CAE6CA}0.01 \\
 &
  \multicolumn{1}{l|}{\textit{Float-32}} &
  \multicolumn{1}{l|}{\cellcolor[HTML]{CAE6CA}80.00\%} &
  \cellcolor[HTML]{CAE6CA}3070.36 &
  \cellcolor[HTML]{CAE6CA}8.15 &
  \cellcolor[HTML]{CAE6CA}0.19 &
  \cellcolor[HTML]{CAE6CA}0.01 \\
 &
  \multicolumn{1}{l|}{\textit{Int-16}} &
  \multicolumn{1}{l|}{\cellcolor[HTML]{CAE6CA}84.00\%} &
  \cellcolor[HTML]{CAE6CA}2989.68 &
  \cellcolor[HTML]{CAE6CA}9.10 &
  \cellcolor[HTML]{CAE6CA}0.21 &
  \cellcolor[HTML]{CAE6CA}0.02 \\
\multirow{-3}{*}{\textbf{Boundary}} &
  \multicolumn{1}{l|}{\textit{Int-8}} &
  \multicolumn{1}{l|}{\cellcolor[HTML]{CAE6CA}74.00\%} &
  \cellcolor[HTML]{CAE6CA}2450.47 &
  \cellcolor[HTML]{CAE6CA}61.34 &
  \cellcolor[HTML]{CAE6CA}1.38 &
  \cellcolor[HTML]{CAE6CA}0.08 \\
\rowcolor[HTML]{EFEFEF} 
\cellcolor[HTML]{EFEFEF} &
  \multicolumn{1}{l|}{\cellcolor[HTML]{EFEFEF}\textit{Float-32}} &
  \multicolumn{1}{l|}{\cellcolor[HTML]{CAE6CA}78.80\%} &
  \cellcolor[HTML]{CAE6CA}2413.11 &
  \cellcolor[HTML]{CAE6CA}36.95 &
  \cellcolor[HTML]{CAE6CA}0.77 &
  \cellcolor[HTML]{CAE6CA}0.03 \\
\rowcolor[HTML]{EFEFEF} 
\cellcolor[HTML]{EFEFEF} &
  \multicolumn{1}{l|}{\cellcolor[HTML]{EFEFEF}\textit{Int-16}} &
  \multicolumn{1}{l|}{\cellcolor[HTML]{CAE6CA}79.00\%} &
  \cellcolor[HTML]{CAE6CA}2284.17 &
  \cellcolor[HTML]{CAE6CA}83.51 &
  \cellcolor[HTML]{CAE6CA}1.72 &
  \cellcolor[HTML]{CAE6CA}0.06 \\
\rowcolor[HTML]{EFEFEF} 
\multirow{-3}{*}{\cellcolor[HTML]{EFEFEF}\textbf{GeoDA}} &
  \multicolumn{1}{l|}{\cellcolor[HTML]{EFEFEF}\textit{Int-8}} &
  \multicolumn{1}{l|}{\cellcolor[HTML]{EFEFEF}37.50\%} &
  2193.12 &
  98.67 &
  2.08 &
  0.08 \\ \hline
\end{tabular}%
}
\end{table}

Pre-processing defenses are evaluated against the adversarial examples produced to attack the vanilla ANNs and QNNs before any defense. We do not craft new adversarial examples for the Defense+ANN or Defense+QNN system as pre-processing defenses are known to disrupt the well-known gradient-masking phenomenon, in which the gradient becomes useless to create an adversarial example, giving a false sense of security \cite{bernhard2019, gorsline2021, lin2019}. The evaluation of Pixel Defend was restricted to CIFAR-10 due to hardware limitations: PixelCNN is a heavy network whose amount of neurons grows proportionally with the size of input images. Therefore, we limited the evaluation scope to the dataset with smaller images. The configuration of Feature Squeezing and Pixel Defend is detailed in Table \ref{tab:defenses_config} (Appendix \ref{ap:attack_defense_config}), while the test accuracy of the ANNs/QNNs enhanced with these defenses is detailed in Table \ref{tab:pre_proc_def_acc}. Tables \ref{tab:results_feature_squeezing} and \ref{tab:pixel_defend} detail the results observed for Feature Squeezing and Pixel Defend, respectively.

\mypara{D4. Feature Squeezing vs. Pixel Defend.} Feature Squeezing can denoise more adversarial examples than Pixel Defend. Feature Squeezing returned a higher adversarial accuracy for almost every attack and bit-width, making it a preferable defense from the robustness point-of-view. Regarding the implementation at the deep edge, the amount of processing that Feature Squeezing requires can effectively be handled by Arm Cortex-M MCUs; however, only on MCUs that feature a compatible DSP or FPU. If the color bit-depth compression can be implemented as a simple shift operation, the spatial smoothing filter is more computationally heavy and would benefit from a DSP-enabled MCU. In contrast, Pixel Defend is not feasible in edge devices powered by Arm Cortex-M MCUs as it imposes a heavy generative model as input pre-processing. The PixelCNN used in this evaluation has more than 2M parameters, which do not fit in MCUs with a few MiB.

\begin{mdframed}[style=remarkstyle]
\mypara{Takeaway \thetakeawyacount \stepcounter{takeawyacount}.} Feature Squeezing can denoise a larger fraction of adversarial examples than Pixel Defend. In contrast to Pixel Defend, Feature Squeezing is portable to edge devices powered by Arm Cortex-M MCUs as long as they feature a DSP.
\end{mdframed}

\mypara{D5. Amount of perturbed pixels ($L_0$ distortion).} Feature Squeezing delivers better results for the CIFAR-10 and VWW models than for Coffee. We argue that the worst performance on Coffee models results from the larger distortion created by the attacks. The Coffee dataset features considerably bigger images than the CIFAR-10 and VWW datasets, naturally leading to attacks to perturbing more pixels. For defenses based on filter smoothing, such as Feature Squeezing, if many pixels are distorted, the filter will struggle to revert the effect of the noise as most neighbor pixels are also very distorted.

However, a closer analysis of Feature Squeezing results on Coffee models reveals that the Boundary attack perturbs more pixels ($L_0$ distortion) than other attacks, yet Feature Squeezing still manages to denoise more adversarial examples than on other attacks. This difference becomes more pronounced when compared to white-box attacks, which perturb almost the same magnitude of pixels as the Boundary attack. The reason behind this lies in the other aspect of Feature Squeezing: the reduction of color bit-width. Acting like quantization, it can remove small perturbations while amplifying larger ones. Since white-box attacks can access exact gradients, they can induce more significant perturbations on more crucial pixels. Conversely, the Boundary attack cannot compute the exact gradient and may overemphasize less relevant pixels. Consequently, while the spatial smoothing filter's performance may be similar across all attacks, the reduction of color bit-width proves more successful in denoising adversarial examples crafted with the Boundary attack than white-box attacks.

\begin{mdframed}[style=remarkstyle]
\mypara{Takeaway \thetakeawyacount \stepcounter{takeawyacount}.} Feature Squeezing is effective against small perturbations, but the spatial smoothing filter starts to fall short as the amount of perturbed pixels increases.
\end{mdframed}

\subsection{Train-Based vs. Pre-Processing Defenses}
From the set of train-based defenses, PGD and Sinkhorn Adversarial Training take the lead in terms of robustness, with a little advantage for the later defense. Nevertheless, we believe that Ensemble Adversarial Training would provide better results if the adversarial examples used during training were generated using more powerful attacks. For the particular case of Pixel Defend, we see no better performance when compared to the much simpler Feature Squeezing. 

Regarding the implementation at the deep edge, train-based defenses are preferable over pre-processing ones as they do not incur additional overhead or interfere with the original model architecture. However, for DSP-enabled MCUs, combining PGD or Sinkhorn Adversarial Training with Feature Squeezing could deliver a robust ML system at the cost of probably insignificant overhead.

\begin{mdframed}[style=remarkstyle]
\mypara{Takeaway \thetakeawyacount \stepcounter{takeawyacount}.} Train-based defenses do not incur additional computation during inference and, therefore, are preferable over pre-processing defenses to be deployed in edge devices powered by Arm Cortex-M MCUs.
\end{mdframed}

\section{Related Work and Gap Analysis}

\begin{table*}[]
\caption{Gap Analysis. Filled-circles ($\CIRCLE$) indicate that a given work empirically evaluates a given takeaway. Empty-circles ($\Circle$) indicate that a given work does not empirically evaluate a given takeaway. N.D. means Non-Defined.}
\label{tab:gap_analysis}
\resizebox{\textwidth}{!}{%
\begin{tabular}{lccccccc}
\hline
\multicolumn{1}{l|}{} &
  \multicolumn{7}{c}{} \\
\multicolumn{1}{l|}{} &
  \multicolumn{7}{c}{\multirow{-2}{*}{\textbf{Work}}} \\ \cline{2-8} 
\multicolumn{1}{l|}{} &
  \multicolumn{1}{c|}{} &
  \multicolumn{1}{c|}{} &
  \multicolumn{1}{c|}{} &
  \multicolumn{1}{c|}{} &
  \multicolumn{1}{c|}{} &
  \multicolumn{1}{c|}{} &
  \cellcolor[HTML]{ECF4FF} \\
\multicolumn{1}{l|}{\multirow{-4}{*}{}} &
  \multicolumn{1}{c|}{\multirow{-2}{*}{\textcite{zhao2018}}} &
  \multicolumn{1}{c|}{\multirow{-2}{*}{\textcite{bernhard2019}}} &
  \multicolumn{1}{c|}{\multirow{-2}{*}{\textcite{lin2019}}} &
  \multicolumn{1}{c|}{\multirow{-2}{*}{\textcite{duncan2020}}} &
  \multicolumn{1}{c|}{\multirow{-2}{*}{\textcite{gorsline2021}}} &
  \multicolumn{1}{c|}{\multirow{-2}{*}{\textcite{song2021}}} &
  \multirow{-2}{*}{\cellcolor[HTML]{ECF4FF}\textbf{Our Work}} \\ \hline
\multicolumn{8}{c}{\cellcolor[HTML]{C0C0C0}\textbf{ATTACKS}} \\ \hline
\multicolumn{1}{l|}{} &
  \multicolumn{1}{c|}{} &
  \multicolumn{1}{c|}{} &
  \multicolumn{1}{c|}{} &
  \multicolumn{1}{c|}{} &
  \multicolumn{1}{c|}{} &
  \multicolumn{1}{c|}{} &
  \cellcolor[HTML]{ECF4FF} \\
\multicolumn{1}{l|}{\multirow{-2}{*}{Point distance to the decision boundary (T1)}} &
  \multicolumn{1}{c|}{\multirow{-2}{*}{$\Circle$}} &
  \multicolumn{1}{c|}{\multirow{-2}{*}{$\Circle$}} &
  \multicolumn{1}{c|}{\multirow{-2}{*}{$\Circle$}} &
  \multicolumn{1}{c|}{\multirow{-2}{*}{$\Circle$}} &
  \multicolumn{1}{c|}{\multirow{-2}{*}{$\CIRCLE$}} &
  \multicolumn{1}{c|}{\multirow{-2}{*}{$\Circle$}} &
  \multirow{-2}{*}{\cellcolor[HTML]{ECF4FF}$\CIRCLE$} \\ \hline
\multicolumn{1}{l|}{} &
  \multicolumn{1}{c|}{} &
  \multicolumn{1}{c|}{} &
  \multicolumn{1}{c|}{} &
  \multicolumn{1}{c|}{} &
  \multicolumn{1}{c|}{} &
  \multicolumn{1}{c|}{} &
  \cellcolor[HTML]{ECF4FF} \\
\multicolumn{1}{l|}{\multirow{-2}{*}{Gradient masking (T2)}} &
  \multicolumn{1}{c|}{\multirow{-2}{*}{$\Circle$}} &
  \multicolumn{1}{c|}{\multirow{-2}{*}{$\CIRCLE$}} &
  \multicolumn{1}{c|}{\multirow{-2}{*}{$\Circle$}} &
  \multicolumn{1}{c|}{\multirow{-2}{*}{$\Circle$}} &
  \multicolumn{1}{c|}{\multirow{-2}{*}{$\Circle$}} &
  \multicolumn{1}{c|}{\multirow{-2}{*}{$\Circle$}} &
  \multirow{-2}{*}{\cellcolor[HTML]{ECF4FF}$\CIRCLE$} \\ \hline
\multicolumn{1}{l|}{} &
  \multicolumn{1}{c|}{} &
  \multicolumn{1}{c|}{} &
  \multicolumn{1}{c|}{} &
  \multicolumn{1}{c|}{} &
  \multicolumn{1}{c|}{} &
  \multicolumn{1}{c|}{} &
  \cellcolor[HTML]{ECF4FF} \\
\multicolumn{1}{l|}{\multirow{-2}{*}{Transferability vs. quantization-shift (T3)}} &
  \multicolumn{1}{c|}{\multirow{-2}{*}{$\CIRCLE$}} &
  \multicolumn{1}{c|}{\multirow{-2}{*}{$\CIRCLE$}} &
  \multicolumn{1}{c|}{\multirow{-2}{*}{$\CIRCLE$}} &
  \multicolumn{1}{c|}{\multirow{-2}{*}{$\CIRCLE$}} &
  \multicolumn{1}{c|}{\multirow{-2}{*}{$\CIRCLE$}} &
  \multicolumn{1}{c|}{\multirow{-2}{*}{$\CIRCLE$}} &
  \multirow{-2}{*}{\cellcolor[HTML]{ECF4FF}$\CIRCLE$} \\ \hline
\multicolumn{1}{l|}{} &
  \multicolumn{1}{c|}{} &
  \multicolumn{1}{c|}{} &
  \multicolumn{1}{c|}{} &
  \multicolumn{1}{c|}{} &
  \multicolumn{1}{c|}{} &
  \multicolumn{1}{c|}{} &
  \cellcolor[HTML]{ECF4FF} \\
\multicolumn{1}{l|}{\multirow{-2}{*}{Gradient misalignment (T4)}} &
  \multicolumn{1}{c|}{\multirow{-2}{*}{$\Circle$}} &
  \multicolumn{1}{c|}{\multirow{-2}{*}{$\CIRCLE$}} &
  \multicolumn{1}{c|}{\multirow{-2}{*}{$\Circle$}} &
  \multicolumn{1}{c|}{\multirow{-2}{*}{$\Circle$}} &
  \multicolumn{1}{c|}{\multirow{-2}{*}{$\Circle$}} &
  \multicolumn{1}{c|}{\multirow{-2}{*}{$\Circle$}} &
  \multirow{-2}{*}{\cellcolor[HTML]{ECF4FF}$\CIRCLE$} \\ \hline
\multicolumn{8}{c}{\cellcolor[HTML]{C0C0C0}\textbf{DEFENSES}} \\ \hline
\multicolumn{1}{l|}{} &
  \multicolumn{1}{c|}{} &
  \multicolumn{1}{c|}{} &
  \multicolumn{1}{c|}{} &
  \multicolumn{1}{c|}{} &
  \multicolumn{1}{c|}{} &
  \multicolumn{1}{c|}{} &
  \cellcolor[HTML]{ECF4FF} \\
\multicolumn{1}{l|}{\multirow{-2}{*}{Point distance to the decision boundary (T5)}} &
  \multicolumn{1}{c|}{\multirow{-2}{*}{$\Circle$}} &
  \multicolumn{1}{c|}{\multirow{-2}{*}{$\Circle$}} &
  \multicolumn{1}{c|}{\multirow{-2}{*}{$\Circle$}} &
  \multicolumn{1}{c|}{\multirow{-2}{*}{$\Circle$}} &
  \multicolumn{1}{c|}{\multirow{-2}{*}{$\Circle$}} &
  \multicolumn{1}{c|}{\multirow{-2}{*}{$\Circle$}} &
  \multirow{-2}{*}{\cellcolor[HTML]{ECF4FF}$\CIRCLE$} \\ \hline
\multicolumn{1}{l|}{} &
  \multicolumn{1}{c|}{} &
  \multicolumn{1}{c|}{} &
  \multicolumn{1}{c|}{} &
  \multicolumn{1}{c|}{} &
  \multicolumn{1}{c|}{} &
  \multicolumn{1}{c|}{} &
  \cellcolor[HTML]{ECF4FF} \\
\multicolumn{1}{l|}{\multirow{-2}{*}{Adversarial example transferability (T6)}} &
  \multicolumn{1}{c|}{\multirow{-2}{*}{$\Circle$}} &
  \multicolumn{1}{c|}{\multirow{-2}{*}{$\Circle$}} &
  \multicolumn{1}{c|}{\multirow{-2}{*}{$\Circle$}} &
  \multicolumn{1}{c|}{\multirow{-2}{*}{$\Circle$}} &
  \multicolumn{1}{c|}{\multirow{-2}{*}{$\Circle$}} &
  \multicolumn{1}{c|}{\multirow{-2}{*}{$\Circle$}} &
  \multirow{-2}{*}{\cellcolor[HTML]{ECF4FF}$\CIRCLE$} \\ \hline
\multicolumn{1}{l|}{} &
  \multicolumn{1}{c|}{} &
  \multicolumn{1}{c|}{} &
  \multicolumn{1}{c|}{} &
  \multicolumn{1}{c|}{} &
  \multicolumn{1}{c|}{} &
  \multicolumn{1}{c|}{} &
  \cellcolor[HTML]{ECF4FF} \\
\multicolumn{1}{l|}{\multirow{-2}{*}{Defense transferability to TinyML (T7)}} &
  \multicolumn{1}{c|}{\multirow{-2}{*}{$\Circle$}} &
  \multicolumn{1}{c|}{\multirow{-2}{*}{$\Circle$}} &
  \multicolumn{1}{c|}{\multirow{-2}{*}{$\Circle$}} &
  \multicolumn{1}{c|}{\multirow{-2}{*}{$\Circle$}} &
  \multicolumn{1}{c|}{\multirow{-2}{*}{$\Circle$}} &
  \multicolumn{1}{c|}{\multirow{-2}{*}{$\Circle$}} &
  \multirow{-2}{*}{\cellcolor[HTML]{ECF4FF}$\CIRCLE$} \\ \hline
\multicolumn{1}{l|}{} &
  \multicolumn{1}{c|}{} &
  \multicolumn{1}{c|}{} &
  \multicolumn{1}{c|}{} &
  \multicolumn{1}{c|}{} &
  \multicolumn{1}{c|}{} &
  \multicolumn{1}{c|}{} &
  \cellcolor[HTML]{ECF4FF} \\
\multicolumn{1}{l|}{\multirow{-2}{*}{Compare relevant pre-processing defenses (T8)}} &
  \multicolumn{1}{c|}{\multirow{-2}{*}{$\Circle$}} &
  \multicolumn{1}{c|}{\multirow{-2}{*}{$\Circle$}} &
  \multicolumn{1}{c|}{\multirow{-2}{*}{$\Circle$}} &
  \multicolumn{1}{c|}{\multirow{-2}{*}{$\Circle$}} &
  \multicolumn{1}{c|}{\multirow{-2}{*}{$\Circle$}} &
  \multicolumn{1}{c|}{\multirow{-2}{*}{$\Circle$}} &
  \multirow{-2}{*}{\cellcolor[HTML]{ECF4FF}$\CIRCLE$} \\ \hline
\multicolumn{1}{l|}{} &
  \multicolumn{1}{c|}{} &
  \multicolumn{1}{c|}{} &
  \multicolumn{1}{c|}{} &
  \multicolumn{1}{c|}{} &
  \multicolumn{1}{c|}{} &
  \multicolumn{1}{c|}{} &
  \cellcolor[HTML]{ECF4FF} \\
\multicolumn{1}{l|}{\multirow{-2}{*}{Distortion of adversarial examples (T9)}} &
  \multicolumn{1}{c|}{\multirow{-2}{*}{$\Circle$}} &
  \multicolumn{1}{c|}{\multirow{-2}{*}{$\Circle$}} &
  \multicolumn{1}{c|}{\multirow{-2}{*}{$\Circle$}} &
  \multicolumn{1}{c|}{\multirow{-2}{*}{$\Circle$}} &
  \multicolumn{1}{c|}{\multirow{-2}{*}{$\Circle$}} &
  \multicolumn{1}{c|}{\multirow{-2}{*}{$\Circle$}} &
  \multirow{-2}{*}{\cellcolor[HTML]{ECF4FF}$\CIRCLE$} \\ \hline
\multicolumn{1}{l|}{} &
  \multicolumn{1}{c|}{} &
  \multicolumn{1}{c|}{} &
  \multicolumn{1}{c|}{} &
  \multicolumn{1}{c|}{} &
  \multicolumn{1}{c|}{} &
  \multicolumn{1}{c|}{} &
  \cellcolor[HTML]{ECF4FF} \\
\multicolumn{1}{l|}{\multirow{-2}{*}{Train-Based vs. Pre-Processing defenses (T10)}} &
  \multicolumn{1}{c|}{\multirow{-2}{*}{$\Circle$}} &
  \multicolumn{1}{c|}{\multirow{-2}{*}{$\Circle$}} &
  \multicolumn{1}{c|}{\multirow{-2}{*}{$\Circle$}} &
  \multicolumn{1}{c|}{\multirow{-2}{*}{$\Circle$}} &
  \multicolumn{1}{c|}{\multirow{-2}{*}{$\Circle$}} &
  \multicolumn{1}{c|}{\multirow{-2}{*}{$\Circle$}} &
  \multirow{-2}{*}{\cellcolor[HTML]{ECF4FF}$\CIRCLE$} \\ \hline
\multicolumn{8}{c}{\cellcolor[HTML]{C0C0C0}\textbf{QUANTIZATION}} \\ \hline
\multicolumn{1}{l|}{Policy} &
  \multicolumn{1}{c|}{N.D.} &
  \multicolumn{1}{c|}{\begin{tabular}[c]{@{}c@{}}Binary Connect\\ Dorefa Net\end{tabular}} &
  \multicolumn{1}{c|}{N.D.} &
  \multicolumn{1}{c|}{TF Lite} &
  \multicolumn{1}{c|}{N.D.} &
  \multicolumn{1}{c|}{N.D.} &
  \cellcolor[HTML]{ECF4FF}TF Lite Micro \\ \hline
\multicolumn{1}{l|}{} &
  \multicolumn{1}{c|}{} &
  \multicolumn{1}{c|}{} &
  \multicolumn{1}{c|}{} &
  \multicolumn{1}{c|}{} &
  \multicolumn{1}{c|}{} &
  \multicolumn{1}{c|}{} &
  \cellcolor[HTML]{ECF4FF} \\
\multicolumn{1}{l|}{\multirow{-2}{*}{Bit-width}} &
  \multicolumn{1}{c|}{\multirow{-2}{*}{{[}4; 28{]}}} &
  \multicolumn{1}{c|}{\multirow{-2}{*}{{[}1; 4{]}}} &
  \multicolumn{1}{c|}{\multirow{-2}{*}{{[}1; 8{]}}} &
  \multicolumn{1}{c|}{\multirow{-2}{*}{8}} &
  \multicolumn{1}{c|}{\multirow{-2}{*}{{[}1; 10{]}}} &
  \multicolumn{1}{c|}{\multirow{-2}{*}{{[}1; 8{]}}} &
  \multirow{-2}{*}{\cellcolor[HTML]{ECF4FF}8; 16} \\ \hline
\multicolumn{1}{l|}{} &
  \multicolumn{1}{c|}{} &
  \multicolumn{1}{c|}{} &
  \multicolumn{1}{c|}{} &
  \multicolumn{1}{c|}{} &
  \multicolumn{1}{c|}{} &
  \multicolumn{1}{c|}{} &
  \cellcolor[HTML]{ECF4FF} \\
\multicolumn{1}{l|}{\multirow{-2}{*}{Quantized Parameters}} &
  \multicolumn{1}{c|}{\multirow{-2}{*}{Weights/Activations}} &
  \multicolumn{1}{c|}{\multirow{-2}{*}{Weights/Activations}} &
  \multicolumn{1}{c|}{\multirow{-2}{*}{Weights/Activations}} &
  \multicolumn{1}{c|}{\multirow{-2}{*}{Weights/Activations}} &
  \multicolumn{1}{c|}{\multirow{-2}{*}{Weights}} &
  \multicolumn{1}{c|}{\multirow{-2}{*}{Weights}} &
  \multirow{-2}{*}{\cellcolor[HTML]{ECF4FF}Weights/Activations} \\ \hline
\end{tabular}%
}
\end{table*}
\subsection{Related Work}
\textcite{zhao2018} pioneered in the evaluation of adversarial example transferability. The authors concluded that QNNs operating in int-8 or lower bit widths show defensive behavior. The authors also envision that clipping activations change the feature space of ANNs, which could offer marginal protection against transferability.

Along the same research line, i.e., adversary example transferability, \textcite{duncan2020} concluded that when the magnitude of the perturbation is small, quantization increases adversarial robustness by eliminating small perturbations; however, when the magnitude of the perturbation is greater than a given threshold, quantization acts as a distortion amplifier (quantization-shift).

\textcite{bernhard2019} were the first to evaluate QNNs robustness on a non-white-box scenario (using gray-box attacks). The authors concluded that QNNs offer only marginal protection as gray-box attacks can still fool the network. Additionally, the authors state that adversarial transferability is sensitive to the quantization-shift phenomenon and the gradient misalignment.

\textcite{gorsline2021} considered the findings of previous works as a baseline and defined a geometric model to evaluate QNN robustness. Through this geometric model, the authors mathematically proved that the mean distance to the decision boundary decreases as the weight precision and the input dimensionality increase. Additionally, the authors introduced the concept of critical attack strength - the attack strength at which the adversarial accuracy of QNNs matches the adversarial accuracy of ANNs.

\textcite{lin2019} followed a different approach from previous works and proposed a training algorithm to tackle quantization-shift. More specifically, the authors proposed a quantization-aware training algorithm that maintains the Lipschitz constant of the network below 1 - the Lipschitz constant measures how much the output changes when the input changes. Additionally, the authors demonstrate that this mechanism can be combined with other defenses, such as Feature Squeezing or Adversarial Training, to enhance the robustness of QNNs.

\textcite{song2021} evaluated the effect of adversarial examples on 8-bit inferior QNNs. They observed that for precisions under 5 bits, adversarially trained QNNs are more vulnerable to adversarial examples than the equivalent ANN. To tackle this issue, they proposed an iteration of PGD Adversarial Training specialized for QNNs.

\subsection{Gap Analysis}
In this section, we put our work in perspective with the works previously described along three different dimensions. Table \ref{tab:gap_analysis} guides our discussion.

\mypara{Attacks.} Information on QNN's adversarial robustness is scattered across multiple works that do not share the same quantization policy. In some cases, this has led to contradictory conclusions \cite{lin2019, duncan2020}. While \textcite{lin2019} states that QNNs are more vulnerable to adversarial examples, \textcite{duncan2020} suggests that QNNs show increased robustness when compared to ANNs. Nevertheless, both of these works only consider the quantization-shift phenomenon (T3) and let out all the other factors that also influence the robustness of a QNN. If we take a closer look at Table \ref{tab:gap_analysis}, we observe that the most complete work till date comes from \textcite{bernhard2019}; however, the authors do not evaluate the point distance to the decision boundary (T1) and do not follow a state-of-the-art quantization policy. Given this background, our work systematizes and empirically evaluates all the conclusions drawn from previous works, under a common quantization policy, which is widely used in TinyML (TF Lite Micro).

Compared to previous research, our empirical evaluation relies on a far more extensive list of attacks. To the best of our knowledge, we are the only ones including a parameter-free attack (AutoAttack), which reduces the chance of bad results due to human error. In the black-box scenario, which is the most realistic, we are the first to include powerful attacks. Previous works have either neglected this scenario or have only included simple attacks that add random noise to images \cite{lin2019, song2021}.

\mypara{Defenses.} The works \cite{lin2019, song2021} are the only ones addressing the topic of defenses against adversarial examples for QNNs; however, in a different approach. \textcite{lin2019} proposes a novel training algorithm to reduce the effect of quantization-shift on QNNs and couples it with adversarial training and Feature Squeezing. \textcite{song2021} proposes an iteration of PGD Adversarial Training targeting weight-quantized models (QNNs whose weights are quantized but activations remain in floating-point) with precision equal to or inferior to 8-bits.

Our work follows a different approach and verifies how current defenses for full-precision ANNs perform applied to QNNs running on MCUs (TinyML). We are the first to empirically evaluate; (i) how quantization interferes with the success of defenses (T5, T7, T9); (ii) how defenses affect the adversarial example transferability from ANNs to QNNs (T6); (iii) which type of pre-processing defenses are preferable to be deployed on MCUs (T8); and (iv) which type of defense (train-based vs. pre-processing) is preferable and in which circumstances (T10).

\mypara{Quantization and TinyML.} As mentioned in Section \ref{sec:scope}, QNNs can be deployed in a myriad of devices, ranging from ultra-low-power MCUs to application-specific integrated circuits (ASICs) and GPUs. However, each target imposes different requirements on the QNN architecture and, above all, quantization policy. Consequently, QNNs targeting GPU devices are closer to conventional ANNs than QNNs targeting TinyML applications. The difference can be expressed in terms of the quantization policy, bit-width, size of kernel filters, and depth of the network. Most of the previous works tend to not be clear about the quantization policy or tend to use policies that enable the approximation of the gradients through Straight-Through Estimator (STE) or Mirror Descent techniques, which are not applicable in TF Lite Micro.

Although the work from \textcite{duncan2020} uses the same quantization policy as our work, the authors only evaluate the adversarial transferability vs. quantization-shift phenomenon (T3). To the best of our knowledge, our work pioneers research on adversarial example robustness for TinyML applications. Our evaluation includes three QNNs, two composing the image applications from the MLPerf Tiny benchmark \cite{tinyml}, and another specifically developed to be executed on Arm Cortex-M MCUs \cite{vita2020}.

\section{Discussion}
Although the effect of adversarial examples on QNNs is a topic that has emerged in the literature over the past few years, the information and findings available are quite scattered and sometimes contradictory at first sight. While some works claim that quantization increases the robustness against adversarial examples \cite{duncan2020}, others claim the opposite \cite{lin2019}. What sometimes these works fail to communicate is the type of devices they target. QNNs targeting GPUs or TPUs usually follow different quantization policies than those targeting MCUs. Our work empirically evaluates all the conclusions drawn from previous works under a common quantization policy, tailored for the TinyML scenario (TF Lite Micro). Furthermore, we are the first to empirically evaluate how current defenses tailored for full-precision ANNs perform when applied to QNNs and adapted to the TinyML environment. Our work unveils findings supported by existing literature (T1, T2, T4), extends existing knowledge (T3), and introduces entirely novel contributions (T5-T10). Although our work targets TinyML applications, only the takeaway T8 is dependent on the architecture of the target processor class. The remaining takeaways are independent of the architecture of the target processor class, i.e., they hold for any ML system following the quantization policy of TF Lite Micro. As a final remark, we envision that developing train-based defenses that can smooth the quantization-shift and gradient misalignment issues is the way to enhance the robustness of QNNs in TinyML applications.

\section{Conclusion}
With this work, we have conducted the most comprehensive empirical evaluation of adversarial example attacks and defenses on QNNs targeting TinyML applications. The study encompassed three QNNs targeting edge ML applications on ten attacks and six popular defenses. With that, we drew a set of observations that provide evidence about (i) the effectiveness of adversarial examples on QNNs, (ii) the impact of bit-width on the adversarial example robustness, (iii) the robustness of defenses in QNNs, and (iv) the deployability of state-of-the-art defenses in resource-constrained MCUs. We are opening all artifacts in the hopes of enabling independent validation of results and encouraging further exploration of the security of QNNs, namely the development of adversarial defenses that meet the requirements of TinyML.

\section*{Acknowledgements}
We would like to thank the anonymous reviewers for their valuable feedback and suggestions. This work is supported by FCT – Fundação para a Ciência e Tecnologia within the R\&D Units Project Scope UIDB/00319/2020. Miguel Costa was supported by FCT grant SFRH/BD/146780/2019.

\printbibliography

\newpage
\onecolumn
\appendices

\section{Attacks/Defenses Configuration} \label{ap:attack_defense_config}
Table \ref{tab:attack_config} details the configuration of the attacks. We configured the attacks such that the accuracy of the target ANN drops below the value of random choice (10\% for CIFAR-10, 50\% for VWW, and 25\% for Coffee). Table \ref{tab:defenses_config} details the configuration of the defenses. For each defense, we experimented with different sets of hyper-parameters around the default values suggested by the defenses' authors and selected the configuration that delivered the best results against the PGD attack.
\begin{table}[h]
\centering
\caption{Attacks configuration}
\label{tab:attack_config}
\resizebox{0.75\textwidth}{!}{%
\begin{tabular}{lllll}
\hline
\textbf{Attack}                                          &                     & \textbf{CIFAR-10}      & \textbf{VWW}           & \textbf{Coffee}        \\ \hline
\rowcolor[HTML]{C0C0C0} 
\multicolumn{5}{c}{\cellcolor[HTML]{C0C0C0}\textit{\textbf{WHITE-BOX}}}                                                                                   \\ \hline
                                                         & Max. Iterations     & 100                    & 100                    & 100                    \\
\multirow{-2}{*}{\textbf{DeepFool}}                      & Overshoot           & 0.008                  & 0.008                  & 0.008                  \\
\rowcolor[HTML]{EFEFEF} 
\cellcolor[HTML]{EFEFEF}                                 & Theta               & 0.08                   & 0.08                   & 0.08                   \\
\rowcolor[HTML]{EFEFEF} 
\multirow{-2}{*}{\cellcolor[HTML]{EFEFEF}\textbf{JSMA}}  & Gamma               & 1.00                   & 1.00                   & 1.00                   \\
                                                         & Max. Iterations     & 10                     & 10                     & 10                     \\
                                                         & Binary Search Steps & 10                     & 10                     & 10                     \\
\multirow{-3}{*}{\textbf{C\&W-$L_2$}}                       & Initial Constant    & 0.01                   & 0.01                   & 0.01                   \\
\rowcolor[HTML]{EFEFEF} 
\cellcolor[HTML]{EFEFEF}                                 & Max. Iterations     & 10                     & 10                     & 10                     \\
\rowcolor[HTML]{EFEFEF} 
\multirow{-2}{*}{\cellcolor[HTML]{EFEFEF}\textbf{C\&W-$L_\infty$}} & Learning Rate    & 0.01 & 0.01 & 0.01 \\
                                                         & Epsilon             & {[}0.0002; 0.0008{]}   & {[}0.0002; 0.0008{]}   & {[}0.0002; 0.0008{]}   \\
                                                         & Step Size           & {[}0.00002; 0.00008{]} & {[}0.00002; 0.00008{]} & {[}0.00002; 0.00008{]} \\
\multirow{-3}{*}{\textbf{PGD}}                           & Restarts            & 20                     & 20                     & 20                     \\
\rowcolor[HTML]{EFEFEF} 
\cellcolor[HTML]{EFEFEF}                                 & Max. Iterations     & 10                     & 10                     & 10                     \\
\rowcolor[HTML]{EFEFEF} 
\cellcolor[HTML]{EFEFEF}                                 & BS Steps            & 10                     & 10                     & 10                     \\
\rowcolor[HTML]{EFEFEF} 
\multirow{-3}{*}{\cellcolor[HTML]{EFEFEF}\textbf{EAD}}   & Initial Constant    & 0.01                   & 0.01                   & 0.01                   \\
\textbf{AutoAttack}                                      & Epsilon             & 0.004                       & 0.004                       & 0.009                       \\ \hline
\rowcolor[HTML]{C0C0C0} 
\multicolumn{5}{c}{\cellcolor[HTML]{C0C0C0}\textit{\textbf{GRAY-BOX}}}                                                                                    \\ \hline
\rowcolor[HTML]{EFEFEF} 
\cellcolor[HTML]{EFEFEF}                                 & Max. Iterations     & 10                     & 50                     & 10                     \\
\rowcolor[HTML]{EFEFEF} 
\cellcolor[HTML]{EFEFEF}                                 & BS Steps            & 5                      & 1                      & 2                      \\
\rowcolor[HTML]{EFEFEF} 
\multirow{-3}{*}{\cellcolor[HTML]{EFEFEF}\textbf{ZOO}}   & Initial Constant    & 0.01                   & 0.01                   & 0.01                   \\
                                                         & Epsilon             & 1                      & 2                      & 2                      \\
                                                         & Max. Iterations     & 2000                   & 1000                   & 2000                    \\
\multirow{-3}{*}{\textbf{SA-$L_2$}}                         & Initial Fraction    & 0.80                   & 0.80                   & 0.80                   \\
\rowcolor[HTML]{EFEFEF} 
\cellcolor[HTML]{EFEFEF}                                 & Epsilon             & 0.015                  & 0.015                  & 0.015                  \\
\rowcolor[HTML]{EFEFEF} 
\cellcolor[HTML]{EFEFEF}                                 & Max. Iterations     & 1000                   & 1000                   & 1000                   \\
\rowcolor[HTML]{EFEFEF} 
\multirow{-3}{*}{\cellcolor[HTML]{EFEFEF}\textbf{SA-$L_\infty$}}   & Initial Fraction & 0.05 & 0.05 & 0.05 \\ \hline
\rowcolor[HTML]{C0C0C0} 
\multicolumn{5}{c}{\cellcolor[HTML]{C0C0C0}\textit{\textbf{BLACK-BOX}}}                                                                                   \\ \hline
                                                         & Epsilon             & 1.0                    & 1.0                    & 1.0                    \\
                                                         & Delta               & 0.10                   & 0.10                   & 0.10                   \\
                                                         & Initial Size        & 100                    & 500                    & 100                    \\
\multirow{-4}{*}{\textbf{BA}}                            & Max. Iterations     & 500                    & 500                    & 500                    \\
\rowcolor[HTML]{EFEFEF} 
\cellcolor[HTML]{EFEFEF}                                 & BS Tolerance        & 0.0001                 & 0.0001                 & 0.01                   \\
\rowcolor[HTML]{EFEFEF} 
\multirow{-2}{*}{\cellcolor[HTML]{EFEFEF}\textbf{GeoDA}} & DCT Dimension       & 75                     & 75                     & 10                     \\ \hline
\end{tabular}%
}
\end{table}
\begin{table}[h]
\centering
\caption{Defenses Configuration}
\label{tab:defenses_config}
\resizebox{0.58\textwidth}{!}{%
\begin{tabular}{lllll}
\hline
\textbf{Defense} &
   &
  \textbf{CIFAR-10} &
  \textbf{VWW} &
  \textbf{Coffee} \\ \hline
\rowcolor[HTML]{C0C0C0} 
\multicolumn{5}{c}{\cellcolor[HTML]{C0C0C0}\textit{\textbf{TRAIN-BASED DEFENSES}}} \\ \hline
 &
  Temperature &
  20 &
  20 &
  20 \\
 &
  Alpha &
  0.1 &
  0.1 &
  0.1 \\
\multirow{-3}{*}{\textbf{\begin{tabular}[c]{@{}l@{}}Defensive\\ Distillation\end{tabular}}} &
  Epochs &
  200 &
  50 &
  25 \\
\rowcolor[HTML]{EFEFEF} 
\cellcolor[HTML]{EFEFEF} &
  Epsilon &
  0.008 &
  0.008 &
  0.008 \\
\rowcolor[HTML]{EFEFEF} 
\cellcolor[HTML]{EFEFEF} &
  Step Size &
  0.0008 &
  0.0008 &
  0.0008 \\
\rowcolor[HTML]{EFEFEF} 
\cellcolor[HTML]{EFEFEF} &
  Iterations &
  10 &
  10 &
  10 \\
\rowcolor[HTML]{EFEFEF} 
\multirow{-4}{*}{\cellcolor[HTML]{EFEFEF}\textbf{\begin{tabular}[c]{@{}l@{}}PGD\\ Adversarial\\ Training ¹\end{tabular}}} &
  Epochs &
  50 &
  50 &
  50 \\
 &
  PGD &
  Same as (1) &
  Same as (1) &
  Same as (1) \\
 &
  FGSM - Epsilon &
  0.008 &
  0.008 &
  0.008 \\
\multirow{-3}{*}{\textbf{\begin{tabular}[c]{@{}l@{}}Ensemble\\ Adversarial\\ Training\end{tabular}}} &
  Epochs &
  50 &
  50 &
  50 \\
\rowcolor[HTML]{EFEFEF} 
\cellcolor[HTML]{EFEFEF} &
  PGD &
  Same as (1) &
  Same as (1) &
  Same as (1) \\
\rowcolor[HTML]{EFEFEF} 
\cellcolor[HTML]{EFEFEF} &
  Sink. Epsilon &
  1.0 &
  \cellcolor[HTML]{EFEFEF}1.0 &
  \cellcolor[HTML]{EFEFEF}1.0 \\
\rowcolor[HTML]{EFEFEF} 
\cellcolor[HTML]{EFEFEF} &
  Sink. Iterations &
  50 &
  50 &
  50 \\
\rowcolor[HTML]{EFEFEF} 
\multirow{-4}{*}{\cellcolor[HTML]{EFEFEF}\textbf{\begin{tabular}[c]{@{}l@{}}Sinkhorn\\ Adversarial\\ Training\end{tabular}}} &
  Sink. Reduction &
  Sum &
  \cellcolor[HTML]{EFEFEF}Sum &
  \cellcolor[HTML]{EFEFEF}Sum \\ \hline
\rowcolor[HTML]{C0C0C0} 
\multicolumn{5}{c}{\cellcolor[HTML]{C0C0C0}\textit{\textbf{INPUT PRE-PROCESSING DEFENSES}}} \\ \hline
 &
  Bit-Depth (ANNs) &
  4 &
  8 &
  8 \\
\multirow{-2}{*}{\textbf{\begin{tabular}[c]{@{}l@{}}Feature\\ Squeezing\end{tabular}}} &
  Window Size &
  2 &
  6 &
  2 \\
\rowcolor[HTML]{EFEFEF} 
\textbf{\begin{tabular}[c]{@{}l@{}}Pixel\\ Defend\end{tabular}} &
  Epsilon &
  38 &
  - &
  - \\ \hline
\end{tabular}%
}
\end{table}

\section{Evaluation of Train-Based Defenses} \label{ap:defenses_full_results}
Tables \ref{tab:results_train_based_defenses1} and \ref{tab:results_train_based_defenses2} represent the full-version of Table \ref{tab:results_train_based_defenses_compact} and detail the results registered for train-based defenses against the most powerful attacks that we evaluate in Section \ref{sec:attacks_eval}. Red highlights samples that affect the accuracy and whose noise is detectable to the human eye. Green highlights samples where accuracy dropped by less than 20\% when compared to the baseline accuracy of the model (Table \ref{tab:models_test_acc}).

\begin{table*}[h]
\caption{Evaluation of train-based defenses}
\label{tab:results_train_based_defenses1}
\resizebox{\textwidth}{!}{%
\begin{tabular}{lllllllllllllllll}
\hline
\textbf{Attack} &
  \multicolumn{1}{l|}{} &
  \multicolumn{5}{c|}{\textbf{CIFAR-10}} &
  \multicolumn{5}{c|}{\textbf{Visual Wake Words}} &
  \multicolumn{5}{c}{\textbf{Coffee Dataset}} \\ \cline{3-17} 
 &
  \multicolumn{1}{l|}{} &
  \multicolumn{1}{c|}{\textit{Acc}} &
  \multicolumn{1}{c}{\textit{$L_0$}} &
  \multicolumn{1}{c}{\textit{$L_1$}} &
  \multicolumn{1}{c}{\textit{$L_2$}} &
  \multicolumn{1}{c|}{\textit{$L_\infty$}} &
  \multicolumn{1}{c|}{\textit{Acc}} &
  \multicolumn{1}{c}{\textit{$L_0$}} &
  \multicolumn{1}{c}{\textit{$L_1$}} &
  \multicolumn{1}{c}{\textit{$L_2$}} &
  \multicolumn{1}{c|}{\textit{$L_\infty$}} &
  \multicolumn{1}{c|}{\textit{Acc}} &
  \multicolumn{1}{c}{\textit{$L_0$}} &
  \multicolumn{1}{c}{\textit{$L_1$}} &
  \multicolumn{1}{c}{\textit{$L_2$}} &
  \multicolumn{1}{c}{\textit{$L_\infty$}} \\ \hline
\rowcolor[HTML]{C0C0C0} 
\multicolumn{17}{c}{\cellcolor[HTML]{C0C0C0}\textit{\textbf{DEFENSIVE DISTILATION}}} \\ \hline
\rowcolor[HTML]{D5D5D5} 
\multicolumn{17}{c}{\cellcolor[HTML]{D5D5D5}\textit{TRANSFERABILITY}} \\ \hline
 &
  \multicolumn{1}{l|}{\textit{Float-32}} &
  \multicolumn{1}{l|}{8.60\%} &
   &
   &
   &
  \multicolumn{1}{l|}{} &
  \multicolumn{1}{l|}{14.30\%} &
   &
   &
   &
  \multicolumn{1}{l|}{} &
  \multicolumn{1}{l|}{3.50\%} &
   &
   &
   &
   \\
 &
  \multicolumn{1}{l|}{\textit{Int-16}} &
  \multicolumn{1}{l|}{47.40\%} &
   &
   &
   &
  \multicolumn{1}{l|}{} &
  \multicolumn{1}{l|}{70.20\%} &
   &
   &
   &
  \multicolumn{1}{l|}{} &
  \multicolumn{1}{l|}{12.10\%} &
   &
   &
   &
   \\
\hline
\end{tabular}%
}
\end{table*}
\begin{table*}[t]
\caption{Evaluation of train-based defenses}
\label{tab:results_train_based_defenses2}
\resizebox{\textwidth}{!}{%
\begin{tabular}{lllllllllllllllll}
\hline
\textbf{Attack} &
  \multicolumn{1}{l|}{} &
  \multicolumn{5}{c|}{\textbf{CIFAR-10}} &
  \multicolumn{5}{c|}{\textbf{Visual Wake Words}} &
  \multicolumn{5}{c}{\textbf{Coffee Dataset}} \\ \cline{3-17} 
 &
  \multicolumn{1}{l|}{} &
  \multicolumn{1}{c|}{\textit{Acc}} &
  \multicolumn{1}{c}{\textit{$L_0$}} &
  \multicolumn{1}{c}{\textit{$L_1$}} &
  \multicolumn{1}{c}{\textit{$L_2$}} &
  \multicolumn{1}{c|}{\textit{$L_\infty$}} &
  \multicolumn{1}{c|}{\textit{Acc}} &
  \multicolumn{1}{c}{\textit{$L_0$}} &
  \multicolumn{1}{c}{\textit{$L_1$}} &
  \multicolumn{1}{c}{\textit{$L_2$}} &
  \multicolumn{1}{c|}{\textit{$L_\infty$}} &
  \multicolumn{1}{c|}{\textit{Acc}} &
  \multicolumn{1}{c}{\textit{$L_0$}} &
  \multicolumn{1}{c}{\textit{$L_1$}} &
  \multicolumn{1}{c}{\textit{$L_2$}} &
  \multicolumn{1}{c}{\textit{$L_\infty$}} \\ \hline
\rowcolor[HTML]{C0C0C0} 
\multicolumn{17}{c}{\cellcolor[HTML]{C0C0C0}\textit{\textbf{SINKHORN ADVERSARIAL TRAINING}}} \\ \hline
\rowcolor[HTML]{D5D5D5} 
\multicolumn{17}{c}{\cellcolor[HTML]{D5D5D5}\textit{TRANSFERABILITY}} \\ \hline
 &
  \multicolumn{1}{l|}{\textit{Float-32}} &
  \multicolumn{1}{l|}{10.90\%} &
   &
   &
   &
  \multicolumn{1}{l|}{} &
  \multicolumn{1}{l|}{16.70\%} &
   &
   &
   &
  \multicolumn{1}{l|}{} &
  \multicolumn{1}{l|}{3.70\%} &
   &
   &
   &
   \\
 &
  \multicolumn{1}{l|}{\textit{Int-16}} &
  \multicolumn{1}{l|}{44.30\%} &
   &
   &
   &
  \multicolumn{1}{l|}{} &
  \multicolumn{1}{l|}{24.90\%} &
   &
   &
   &
  \multicolumn{1}{l|}{} &
  \multicolumn{1}{l|}{14.40\%} &
   &
   &
   &
   \\
\hline
  \end{tabular}%
}
\end{table*}

\end{document}